%% file: main.tex
\documentclass[conference]{IEEEtran}
\IEEEoverridecommandlockouts

\usepackage{times}

\usepackage[numbers]{natbib}
\usepackage{multicol}
\usepackage[bookmarks=true]{hyperref}
\usepackage{graphicx}
\usepackage{subcaption}
\usepackage{float}
\usepackage{amsmath,amssymb}
\usepackage{bm}
\usepackage{xspace}
\newcommand{\ie}{\emph{i.e.,}\xspace}
\newcommand{\eg}{\emph{e.g.,}\xspace}
\usepackage{caption}

\DeclareMathOperator{\SIM}{SIM}
\DeclareMathOperator{\SE}{SE}
\usepackage{amsthm}
\usepackage{mathtools}
\providecommand{\linkToPdf}[1]{\url{#1}} % or {} to hide it
\providecommand{\award}[1]{}             % hides award text; or format it as you like

\newcommand{\best}[1]{\cellcolor{green!60}{{#1}}}
\newcommand{\second}[1]{\cellcolor{yellow!60}{#1}}
\usepackage{makecell}
\newcommand{\myParagraph}[1]{{\bf #1.}\xspace}
\usepackage{xcolor}
\usepackage{tikz}
\usepackage{booktabs}
\usepackage{multirow}
\usepackage{colortbl}
\usepackage{array}
\usetikzlibrary{calc}
\usepackage[framemethod=tikz]{mdframed}
\include{preamble_symbols}

\usepackage{adjustbox}
\usepackage{cuted}
\IEEEaftertitletext{\vspace{-4.8\baselineskip}}
\usepackage{capt-of}

\newtheorem{remark}{Remark}
\newtheoremstyle{runindef}%
  {3pt}{3pt}% space above/below
  {\normalfont}{}% body font, indent
  {\bfseries}{}% head font, punctuation
  {.75em}% space after head
  {\thmname{#1}\thmnumber{ #2}\thmnote{ \normalfont(#3)}}% head spec

\theoremstyle{runindef}
\newtheorem{definition}{Definition}
\newmdenv[
  roundcorner=8pt,
  linewidth=1.2pt,
  linecolor=black!55,
  backgroundcolor=black!10,
  innerleftmargin=18pt,
  innerrightmargin=18pt,
  innertopmargin=4pt,
  innerbottommargin=14pt,
  frametitle={\bfseries Summary},
  frametitlerule=false,
  frametitlealignment=\centering,
]{summarybox}

\newmdenv[
  roundcorner=8pt,
  linewidth=1.2pt,
  linecolor=black!55,
  backgroundcolor=black!10,
  innerleftmargin=18pt,
  innerrightmargin=18pt,
  innertopmargin=4pt,
  innerbottommargin=14pt,
  frametitle={\bfseries },
  frametitlerule=false,
  frametitlealignment=\centering,
]{plainbox}

\newmdenv[
  roundcorner=8pt,
  linewidth=1.2pt,
  linecolor=black!55,
  backgroundcolor=black!10,
  innerleftmargin=18pt,
  innerrightmargin=18pt,
  innertopmargin=4pt,
  innerbottommargin=14pt,
  frametitle={\bfseries Takeaway},
  frametitlerule=false,
  frametitlealignment=\centering,
]{takeawaybox}

\usepackage{booktabs}

\begin{document}

\title{Picasso: Holistic Scene Reconstruction \\ with Physics-Constrained Sampling}

% You will get a Paper-ID when submitting a pdf file to the conference system
\author{
Xihang Yu$^{\dagger}$\quad \thanks{This work was partially supported by Symbotic, the ONR RAPID program, Carlone’s
NSF CAREER award, and AFOSR “Certifiable and Self-Supervised Category-Level Tracking” Program.}
Rajat Talak$^{\ddagger}$\quad
Lorenzo Shaikewitz$^{\dagger}$\quad \thanks{L. Shaikewitz is supported by an NSF graduate research fellowship.}
Luca Carlone$^{\dagger}$\thanks{L. Carlone holds concurrent appointments as a faculty at the Massachusetts Institute of Technology and as an Amazon Scholar. This paper describes work performed at MIT and is not associated with Amazon.}
\\[2mm]
{\small $^{\dagger}$ Massachusetts Institute of Technology, Cambridge, MA, 02139, USA}\\
{\small $^{\ddagger}$ National University of Singapore, Singapore 117583}\\
{\small Emails: \texttt{\{jimmyyu, lorenzos, lcarlone\}@mit.edu}}\\
{\small Email: \texttt{talak@nus.edu.sg}}
}
%\author{\authorblockN{Michael Shell}
%\authorblockA{School of Electrical and\\Computer Engineering\\
%Georgia Institute of Technology\\
%Atlanta, Georgia 30332--0250\\
%Email: mshell@ece.gatech.edu}
%\and
%\authorblockA{Twentieth Century Fox\\
%\and
%\authorblockN{James Kirk\\ and Montgomery Scott}
%\authorblockA{Starfleet Academy\\
%San Francisco, California 96678-2391\\

% avoiding spaces at the end of the author lines is not a problem with
% conference papers because we don't use \thanks or \IEEEmembership

%Homer Simpson\authorrefmark{2},
%Montgomery Scott\authorrefmark{3} and
%Eldon Tyrell\authorrefmark{4}}
%Telephone: (800) 555--1212, Fax: (888) 555--1212}
%\authorblockA{\authorrefmark{4}Tyrell Inc., 123 Replicant Street, Los Angeles, California 90210--4321}}

\makeatletter

% \twocolumn[{%
% \begin{@twocolumnfalse}

% \maketitle

% % \begin{center}
% %     \includegraphics[width=0.9\textwidth]{figures/picasso_dataset.pdf}
% %     \captionof{figure}{We propose \emph{Picasso}, an approach to build multi-object scene reconstructions by accounting for object geometry, non-penetration, and physics (\ie objects should be in a stable equilibrium for the scene to be static). 
% %     % This is achieved by a sampling-based approach that reasons over multi-object interactions in a tractable way by leveraging the fact only a subset of objects are in mutual contact in the scene, which helps breaking down the complexity of reasoning over mutual interactions. 
% %     We also release the \emph{Picasso dataset}: a collection of 10 contact-rich real-world scenes we use to test physical plausibility of scene reconstructions.
% %     The figure shows the digital twins generated from the 10 real-world scenes using ground-truth pose annotations.
% %     % builds on existing approaches for object pose and shape estimation and refines 
% %     %Picasso Dataset. It contains 10 contact-rich real-world scenes. Shown is the digital twin generated in simulation using pose annotations from real images.
% %     }
% %     \label{fig:dataset}
% % \end{center}
% % \vspace{0.5em}

% \end{@twocolumnfalse}
% }]

\maketitle

\begin{strip}
  \centering
  \includegraphics[width=0.95\textwidth]{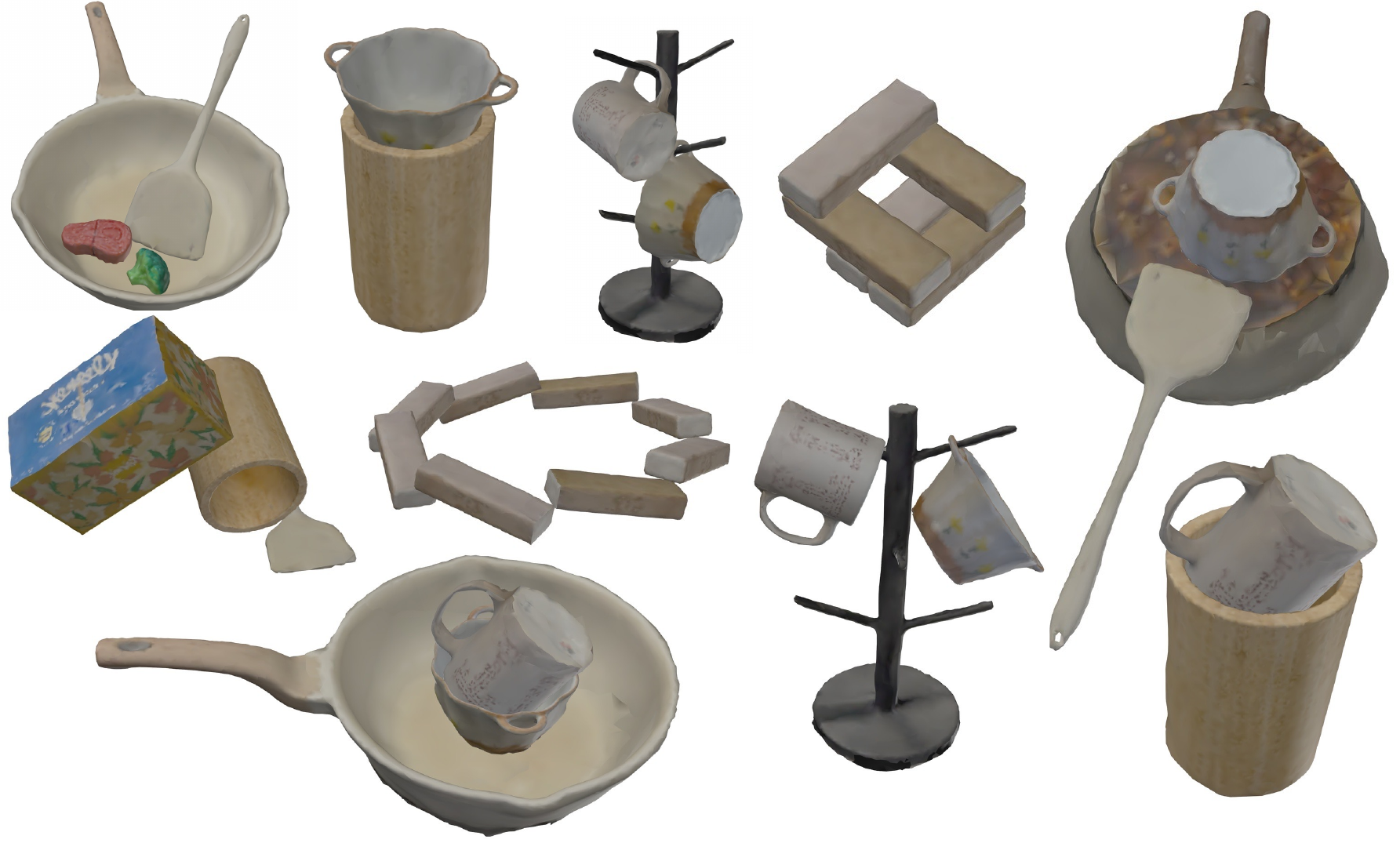}
    \captionof{figure}{We propose \emph{Picasso}, an approach to build multi-object scene reconstructions by accounting for object geometry, non-penetration, and physics (\ie objects should be in a stable equilibrium for the scene to be static). 
    % This is achieved by a sampling-based approach that reasons over multi-object interactions in a tractable way by leveraging the fact only a subset of objects are in mutual contact in the scene, which helps breaking down the complexity of reasoning over mutual interactions. 
    We also release the \emph{Picasso dataset}: a collection of 10 contact-rich real-world scenes we use to test physical plausibility of scene reconstructions.
    The figure shows the digital twins generated from the 10 real-world scenes using ground-truth pose annotations.
    % builds on existing approaches for object pose and shape estimation and refines 
    %Picasso Dataset. It contains 10 contact-rich real-world scenes. Shown is the digital twin generated in simulation using pose annotations from real images.
    }
  \label{fig:dataset}
\end{strip}

\makeatother

\input{sections/abstract}

\input{sections/introduction}
\input{sections/relatedwork}
% \input{sections/prelim}
% \input{sections/problem_formulation}
\input{sections/theoretical_background}
\input{sections/method}
\input{sections/dataset}
\input{sections/experiment}
\input{sections/limitations_future_work}
\input{sections/conclusion}
% \section*{Acknowledgments}
\IEEEpeerreviewmaketitle

%% Use plainnat to work nicely with natbib. 

\bibliographystyle{plainnat}
\bibliography{references/refs,references/myRefs}

\clearpage
\input{sections/appendix.tex}

\end{document}

%% file: preamble_symbols.tex
\newcommand{\calA}{{\cal A}}
\newcommand{\calB}{{\cal B}}

\newcommand{\calE}{{\cal E}}
\newcommand{\calF}{{\cal F}}
\newcommand{\calG}{{\cal G}}

\newcommand{\calP}{{\cal P}}

\newcommand{\calV}{{\cal V}}

\newcommand{\M}[1]{{\bm #1}} % Face for matrices
\renewcommand{\boldsymbol}[1]{{\bm #1}}
\newcommand{\tran}{^{\mathsf{T}}}

\newcommand{\eye}{{\mathbf I}}

\newcommand{\MD}{\M{D}}

\newcommand{\MM}{\M{M}}

\newcommand{\MR}{\M{R}}
\newcommand{\MS}{\M{S}}
\newcommand{\MI}{\M{I}}

\newcommand{\MT}{\M{T}}

\newcommand{\va}{\boldsymbol{a}} 
 
\newcommand{\vb}{\boldsymbol{b}}

\newcommand{\vv}{\boldsymbol{v}}
\newcommand{\vt}{\boldsymbol{t}}

\newcommand{\vtheta}{\boldsymbol{\theta}}

%% file: sections/abstract.tex
%!TEX root = ../main.tex

\begin{abstract}
In the presence of occlusions and measurement noise, 
geometrically accurate scene reconstructions---which fit the sensor data---can still be \emph{physically incorrect}. 
For instance, when estimating the poses and shapes of objects in the scene and importing the resulting estimates into a simulator, small errors might translate to implausible configurations including object interpenetration or unstable equilibrium. 
This makes it difficult to predict the dynamic behavior of the scene using a digital twin, an important step in simulation-based planning and control of contact-rich behaviors.
In this paper, we posit that object pose and shape estimation requires reasoning holistically over the scene (instead of reasoning about each object in isolation), accounting for object interactions and physical plausibility.
Towards this goal, our first contribution is \emph{Picasso}, 
a physics-constrained reconstruction pipeline that builds multi-object scene reconstructions by considering geometry, non-penetration, and physics.
Picasso relies on a fast rejection sampling method that reasons over multi-object interactions, leveraging an inferred object contact graph to guide samples.
Second, we propose the \emph{Picasso dataset}, a collection of 10 contact-rich real-world scenes with ground truth annotations, 
as well as a metric to quantify physical plausibility, which we open-source as part of our benchmark. 
Finally, we provide an extensive evaluation of Picasso on our newly introduced dataset and on the YCB-V dataset, and show it largely outperforms the state of the art while providing reconstructions that are both physically plausible and more aligned with human intuition.
\end{abstract}

%% file: sections/introduction.tex
%!TEX root = ../main.tex

\section{Introduction}
\label{sec:introduction}

Imagine playing Jenga or building a toy castle with wooden blocks. To decide how to remove or add blocks, we would form a mental picture of the structure in front of us, and reason over relations between blocks (for instance, ``block A supports blocks B and C'').
This allows us to ponder different actions (``if I remove block A, two blocks will fall''); in other words, such a mental model serves as a digital twin (or, in modern terms, as an explicit \emph{world model}) we can use to predict how the environment would respond to our actions.
While being effortless for humans, the task of building such a world model is still challenging for robots, and requires forming a holistic scene understanding that accounts for both visual observations and physical plausibility.
At the same time, unlocking this capability for our robots would enable contact-rich manipulation and reasoning in cluttered scenes, 
possibly  
beyond the reach of 
%where the complex reasoning might still be elusive for 
current vision-language-action models. 
%to develop better real-to-sim-to-real approaches for contact-rich manipulation~\cite{FIND REF}, which build the world model from sensor observations, decide the best course of action using such model, and applies the resulting actions in the real world.
% Our visual system is designed with physics intuition. Imagine yourself playing a Super Mario Game for the first time: Soon, you know how to jump, run, following off a ledge. Soon, you learned how to hit blocks from below which can send items flying or bounce the block, or escape unique traps like swinging balls, grinders, dropping pillars. Although the physics of Super Mario may be very different from real physics, but the physics behind the scene is similar to the real one. That is the reason why you don't necessarily know the "physics" inside of the game physics engine, but you adapt to the physics of the game very quickly. 

\begin{figure}[t]
\centering
  \includegraphics[width=\columnwidth]{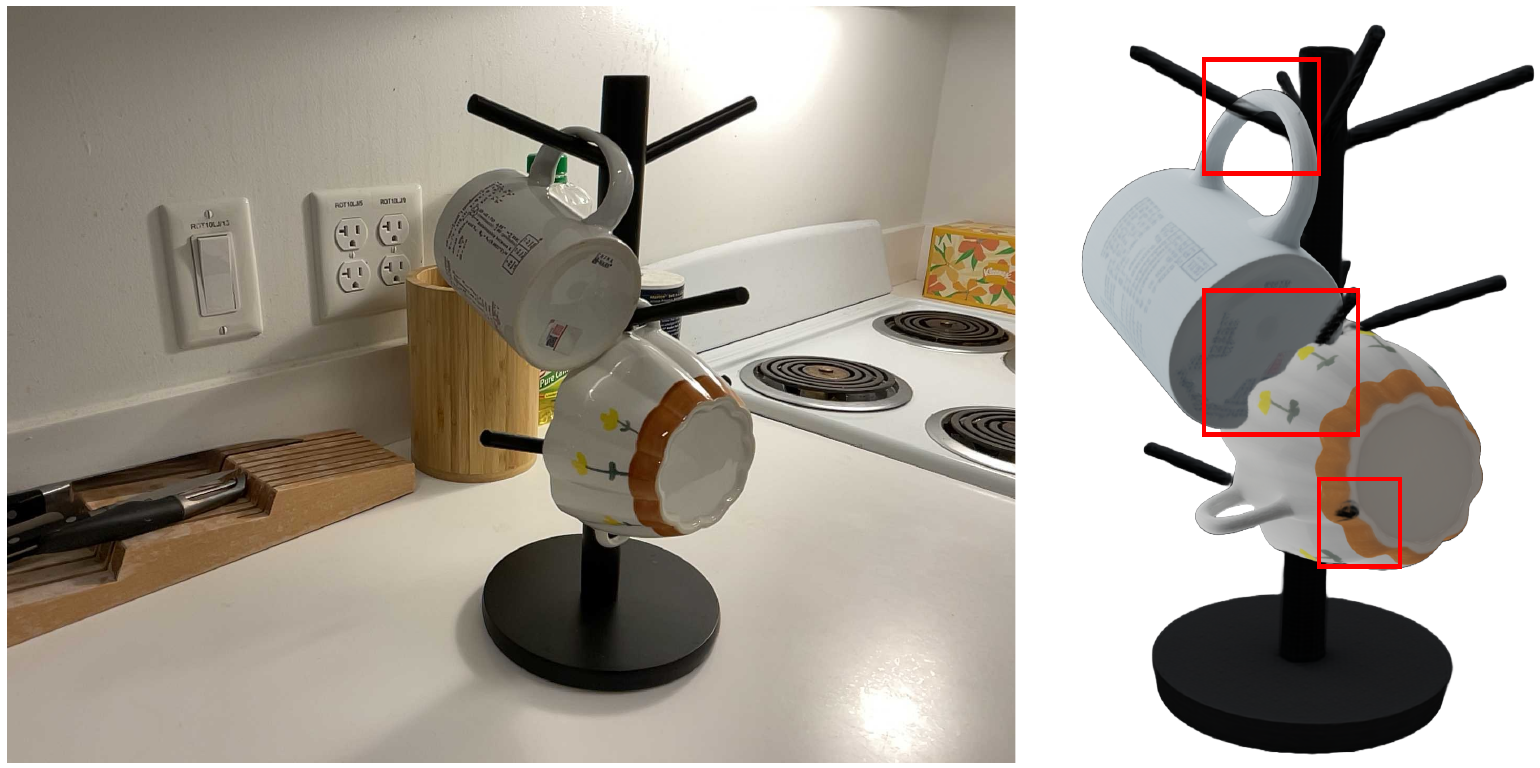}
\caption{An example illustrating that a 3D scene reconstruction from SAM3D~\cite{chen25arxiv-sam} can be physically implausible. \textbf{Left}: The original image. \textbf{Right}: The reconstruction exhibits multiple penetrations, highlighted in the red boxes.
\label{fig:teaser}
\vspace{-5mm}}
\end{figure}

%% limitations of sota
Despite the fast-paced progress in computer vision, graphics, and robotics, only a few works deal with physics-aware explicit world modeling~\cite{puyin25arxiv-quantiphy}. Foundation models for 3D reconstruction~\cite{wang25cvpr-vggt} build a dense 3D reconstruction of a scene, but do not reason about interactions among objects. Work on object understanding, including recent foundation 3D vision models~\cite{chen25arxiv-sam}, most commonly focus on estimating pose and shape of objects, but process objects in isolation; this leads to artifacts where even fairly accurate pose estimates lead to physically implausible configurations that are immediately recognizable as incorrect by a human (Fig~\ref{fig:teaser}). 
To incorporate physics constraints, a common strategy is to design physics-guided losses, \eg discouraging copenetration and encouraging stable contact---and optimize them with gradient-based methods~\cite{agnew21corl-amodal,malenicky25arxiv-physpose,yao25tg-cast}. However, this approach is prone to converge to local minima, which are physically plausible but geometrically incorrect configurations. 
 %For instance, when two objects copenetrate, the gradient of the loss can push the objects away from their actual configuration.
 % We argue, however, that first-order optimization can produce physically plausible but incorrect corrections: when two objects interpenetrate, the gradient may separate them along a locally convenient direction that resolves the loss while pushing the objects away from their true configuration. 
 An alternative is to integrate a differentiable physics simulator to generate corrective signals (\eg~\cite{jiang25arxiv-phystwin, ni24neurips-phyrecon}); however, simulation-based supervision is often brittle in practice due to modeling inaccuracies and numerical noise in the simulation engine~\cite{yao25tg-cast}.

In this paper, we posit that object pose and shape estimation requires reasoning \emph{holistically} over the scene (instead of reasoning about each object in isolation) and accounting for object interactions and physical plausibility. % of the estimated scene.
Towards this goal, we propose a model-based approach that retrofits existing object pose and shape estimators (\eg SAM3D~\cite{chen25arxiv-sam}) to make their estimates more physically plausible.
%, but show it complements well state-of-the-art learning-based methods, such as SAM3D~\cite{chen25arxiv-sam}. 

In particular, we propose \emph{Picasso} (Section~\ref{sec:method}), 
a physics-constrained reconstruction pipeline that builds multi-object scene reconstructions by accounting for geometry, non-penetration, and contact. 
Instead of directly optimizing a physics-based loss, Picasso relies on fast rejection sampling. 
This has two advantages. First, by treating physics as constraints rather than as additional penalty terms, we avoid reshaping the objective with competing losses that may introduce extra local minima. Second, rejection sampling encourages global exploration of the state space, which further reduces the risk of being stuck in local minima. 
To make the sampling tractable, we sample poses that respect which objects are in contact. This reduces the complexity of reasoning over hypothetical interactions and reduces the dimensionality of the sampling space. 
% TODO: would be better to link to github, and have data linked from github.

Our second contribution is the \emph{Picasso dataset} (Section~\ref{sec:dataset}), a collection of 10 contact-rich real-world scenes with ground truth annotations, 
as well as novel metrics to quantify physical plausibility, which we open-source with our benchmark. 
The Picasso dataset is visualized in Fig.~\ref{fig:dataset} (the figure shows the ground-truth digital twin of each scene, while sample images are given in Appendix \ref{app:dataset}) and includes multiple contact-rich scenes found in domestic environments (\eg a pile of plates and pans in a sink, or a set of Jenga blocks).  

Our final contribution is an \emph{extensive evaluation} of Picasso (Section~\ref{sec:experiments}), on both our newly introduced dataset and on the YCB-V dataset \cite{Xiang17rss-posecnn}. We show that the proposed approach (i) can easily retrofit modern object pose and shape estimators, (ii) outperforms state-of-the-art methods in terms of pose accuracy 
and physical plausibility,
% but more importantly it leads to more physically plausible scene reconstructions, 
and (iii) the resulting reconstructions are more aligned with human intuition. Our code and dataset are available online\footnote{\url{https://github.com/MIT-SPARK/Picasso}}.

%  largely outperforms the state of the art while providing reconstructions that are both physically plausible and more aligned with human intuition.
% In this project, we propose instead to tackle hard physics constraints via global sampling for physics-constrained pose estimation. Moreover, while simulators may be insufficiently faithful for scene reconstruction correction, we argue they are well-suited for evaluation and benchmarking. We further introduce a human physics plausibility judgments to validate a simulator-based metric. Building on these ideas, we present \textbf{Picasso}, a physics-constrained efficient scene reconstruction framework with a GPU-optimized parallel sampling.
% The key contributions of this project are threefold: (1) a system that enforces physics constraints while reconstructing scenes efficiently; (2) a physically grounded, simulator-based evaluation metric and a new contact-rich dataset, Picasso; and (3) extensive experiments spanning closed-set and open-set settings, along with human-alignment studies.

%% file: sections/relatedwork.tex
%!TEX root = ../main.tex

\section{Related Work}
\label{sec:related_work}

\myParagraph{Object Pose and Shape Estimation}
Object pose and shape estimation involves recovering the pose and shape of an object, typically from RGB or RGB-D observations. Existing methods can be classified into three categories: \emph{instance-level}, \emph{category-level}, and \emph{category-agnostic}. Instance-level approaches assume the object shape is known and focus on pose estimation~\cite{Xiang17rss-posecnn, Talak23tro-c3po,Shi23rss-ensemble, Wen24cvpr-FoundationPose}. In contrast, category-level methods estimate pose and shape of objects within the same object category (\eg cars), without requiring instance-specific CAD models. These approaches often learn to model shape deformations or normalized coordinate representations to capture intra-class variations~\cite{Pavlakos17icra-semanticKeypoints,Wang19-normalizedCoordinate,Tian20eccv-SPD,fu22neurips-category,Shi25cvpr-CRISP, Yu25arxiv-boxPoseShape}.
More recently, category-agnostic %(and usually open-set) 
approaches have attracted increasing attention \cite{iwase24eccv-zero, iwase25cvpr-zerograsp, agarwal25ral-scenecomplete, huang2025cupid}. Compared to category-level methods, which often rely on curated category annotations, 
%and substantial manual labeling and therefore is often studied in a closed-set setting, 
category-agnostic approaches estimate object pose and shape without assuming knowledge of the object category, leading to 
more scalable and broadly applicable methods~\cite{deitke23neurips-objaverse}. Most approaches in this area (\eg~\cite{Wen23cvpr-bundlesdf}) rely on multiple reference images to reconstruct object shape. In contrast, recent work~\cite{chen25arxiv-sam} demonstrates single-image shape and pose estimation.

\myParagraph{Physics-informed Scene Understanding}
% We have been discussing physics intuition, but how can we inject physics intuition into the system? 
% The question of how to inject physical reasoning into robotics and scene undertstanding has been investigated in several works.
% This question has long been a big challenge in scene understanding. 
% What lies in the center of this area is how to inject physics priors into the pipeline. 
Existing methods to inject physical reasoning into scene understanding can be broadly classified into three classes. First, optimization-based methods impose physics-inspired constraints and solve for physically plausible states. Zhang et al.~\cite{zhang22icra-non} focus on multi-object pose estimation by enforcing hard non-penetration constraints and solving the resulting problems using semi-infinite programming. Methods like PhysPose \cite{agnew21corl-amodal, malenicky25arxiv-physpose, yao25tg-cast} instead define differentiable losses based on physics constraints like gravity and non-penetration and optimize the losses with gradient descent. Vysics and follow-up works \cite{bianchini25arxiv-vysics, zhu25arxiv-object} assume a rigid contact model and solve for contact parameters, which are then used for shape estimation. Second, simulation-based approaches rollout dynamics to optimize physics parameters, see PhyRecon~\cite{ni24neurips-phyrecon}, TwinTrack~\cite{yang25arxiv-twintrack}, HoloScene~\cite{xia2025holoscene} and PhysTwin~\cite{jiang25arxiv-phystwin}. Third, sampling approaches formulate physical plausibility in probabilistic terms and perform inference via sampling. Early work~\cite{Zheng13cvpr-sceneUnderstanding} measures plausibility through changes in system energy and optimizes for stability with Markov Chain Monte Carlo (MCMC). Another work, 3DP3~\cite{Gothoskar21arxiv-3dp3}, models flush contact between objects (\ie forces two contact faces to be coplanar with zero offset) and uses MCMC to jointly infer object shapes, poses, and a scene graph. Follow-up work~\cite{gothoskar23arxiv-bayes3d} uses Bayesian inference for objects in flush contact. Our approach is sampling-based but can handle arbitrary contact types (Section~\ref{sec:constraints}) and accounts for co-penetration.

%% file: sections/theoretical_background.tex
% \section{Theoretical Background}
% \label{sec:theoretical_back}

%% file: sections/method.tex
%!TEX root = ../main.tex

\section{Picasso}
\label{sec:method}

\myParagraph{Problem Formulation and Approach}
\label{sec:problem_formulation}
Consider a scene with $N$ objects. 
% \LC{what does this mean?:  Without loss of generality, we treat environmental object separately and denote it as $r$.}  
We are given an RGB image $\MI$, depth map $\MD$, along with object masks $\MM_i, i\in[N]$.
For each object $i\in[N]$ in the scene, our goal is to estimate its scale, translation, and rotation represented in the camera frame, namely $\MT_i=(s_i, \MR_i, \vt_i) \in \SIM(3)$, 
as well as the object shape $\MS_i$. 
 % be the unknown (to-be-computed) scale, rotation, and translation and $\MS_i, i\in[N]$ be the shape of objects.
We formulate this as maximum likelihood estimation (MLE):
\begin{equation}
\underset{\{\MT_i,\MS_i\,\}_{i=1}^{N} \in \calF}{\arg\max}\;
p\!\left(\{\MT_i\;,\;\MS_i\}_{i=1}^{N} \mid \MI, \MD, \{\MM_i\}_{i=1}^{N}\right),
\label{eq:registration}
\end{equation}
% where $\MS_{[N]} \doteq \MS_1,\ldots,\MS_N$, $\vt_{[N]} \doteq \vt_1,\ldots,\vt_N$, 
where $\calF$ is the set of physically plausible pose and shape estimates. Problem~\eqref{eq:registration} finds the most likely object configuration given the measurements, while satisfying physical plausibility. 
% With slight abuse of notation, we treat the background (\eg the table the objects are laying on) as an object itself, with the only difference being that we assume its (planar) shape to be known and we only estimate the plane parameters instead of a full $\SIM(3)$ transformation.
% We formalize the measurement likelihood and feasible set below.
%  The reconstruction problem is modeled as a conditional distribution $p\!\left(\MS_1,\ldots,\MS_N,\;\vt_1,\ldots,\vt_N \mid \MI, \MD, \MM_1,\ldots,\MM_N\right)$. For notational convenience, we abbreviate this as $p\!\left(\MS_{[N]},\;\vt_{[N]} \mid \MI, \MD, \MM_{[N]}\right)$. We are interested in the following problem:
% \begin{equation}
% \underset{\{\MS_i,\,\vt_i\}_{i=1}^{N}}{\arg\max}\;
% p\!\left(\MS_{[N]},\;\vt_{[N]} \mid \MI, \MD, \MM_{[N]}\right).
% \label{eq:registration}
% \end{equation}
% Generally, the probabilistic model captures not only geometric validity but also physical plausibility.

\begin{figure}[t]
\centering
  \includegraphics[width=\columnwidth]{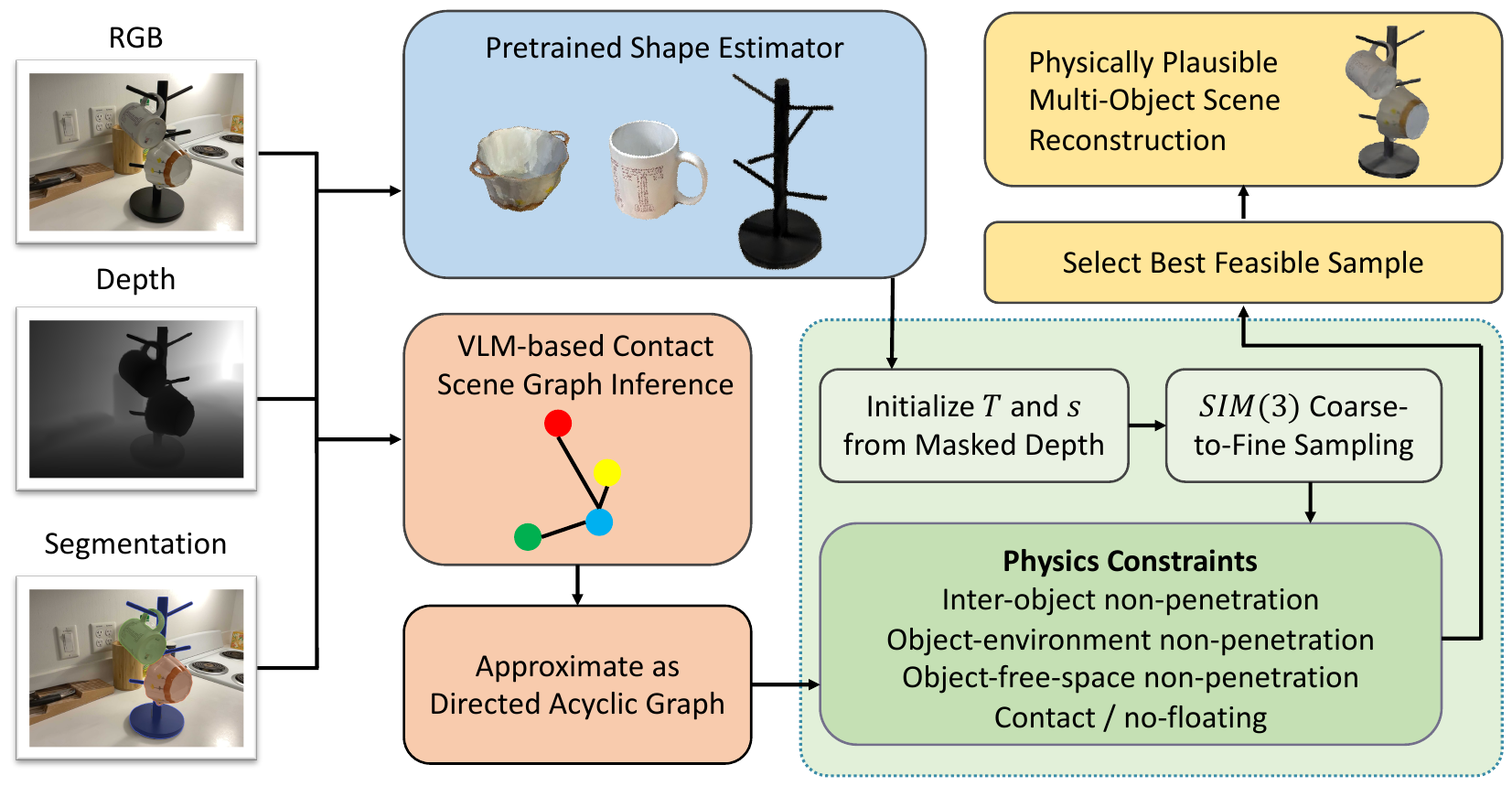}
\caption{System architecture of Picasso. A VLM is queried to infer the contact scene graph, and a pretrained shape estimator provides shape priors. A physics-constrained sampling solver then optimizes object poses by reasoning over multi-object relationships encoded in the contact graph.
\label{fig:block_giagram}
\vspace{-6mm}}
\end{figure}

% \myParagraph{Approach Overview}
% \LC{add approach overview}
Next, we derive expressions for the measurement likelihood (Section~\ref{sec:likelihood}) and constraints 
(Section~\ref{sec:constraints}) in eq.~\eqref{eq:registration}, leading to 
the optimization problem in (Section~\ref{sec:optproblem}) eq.~\eqref{final_problem}.
Then we show how to approximate the problem as a set of lower-dimensional subproblems (one per object), using the notion of \emph{contact scene graph}
 (Section~\ref{sec:simplify_form}). Finally, we solve each subproblem via rejection sampling
 (Section~\ref{sec:rejSampling}).
Figure~\ref{fig:block_giagram} shows a block diagram of the proposed approach.
% (intuitively, rather than sampling all possible object configurations, we only need to account for adjacent objects).

%%%%%%%%%%%%%%%%%%%%%%%%%%%%%%%%%%%%%%%%%%%%%%%%%%%%%%%%%%%%%%%%%%%%%%
%%%%%%%%%%%%%%%%%%%%%%%%%%%%%%%%%%%%%%%%%%%%%%%%%%%%%%%%%%%%%%%%%%%%%%
\subsection{Measurement Likelihood} %Point Cloud Registration}
\label{sec:likelihood}

We begin by defining the measurement likelihood objective in eq.~\eqref{eq:registration}.
% (Section~\ref{sec:likelihood})  
Observe that we can factor the probability in the objective of eq.~\eqref{eq:registration} as:
% In this section, we introduce Picasso, a shape and pose estimation framework with grounded physics constraints. Come back the original problem discussed in Section \ref{sec:problem_formulation}. We can rewrite shape and pose estimation using Bayes Rule:
% \vspace{-1mm}
\begin{equation}
\begin{split}
p\!\left(\{\MT_i\;,\;\MS_i\}_{i=1}^{N} \mid \MI, \MD, \{\MM_i\}_{i=1}^{N}\right)
= \\
p\!\left(\{\MT_i\}_{i=1}^{N} \mid \{\MS_i\}_{i=1}^{N}, \MI, \MD, \{\MM_i\}_{i=1}^{N}\right) \times \\
p\!\left(\{\MS_i\}_{i=1}^{N} \mid \MI, \MD, \{\MM_i\}_{i=1}^{N}\right)
% &p\!\left(\MS_{[N]},\;\MT_{[N]} \mid \MI, \MD, \MM_{[N]}\right) \\
% \propto\; &p\!\left(\MT_{[N]} \mid \MI, \MD, \MM_{[N]}, \MS_{[N]}\right)
% \cdot\; p\!\left(\MS_{[N]} \mid \MI, \MD, \MM_{[N]}\right)
\end{split}
\label{eq:registration2}
\end{equation}
% \vspace{-1mm}
On the left side of \eqref{eq:registration2}, the first term captures pose estimation given the shape, and the second term is shape estimation.

% second term captures shape estimation, while the first term captures pose estimation given the shape.

\myParagraph{Shape Likelihood} Instead of analytically expanding the shape likelihood, 
% We do not derive an analytical expression of the shape likelihood $p\!\left(\{\MS_i\}_{i=1}^{N} \mid \MI, \MD, \{\MM_i\}_{i=1}^{N}\right)$. 
we assume the availability of a network, \eg~\cite{chen25arxiv-sam}, that performs shape prediction given sensor data. 
We can deal with both the case in which the network returns a single shape estimate as in~\cite{chen25arxiv-sam}, as well as the 
case where the network returns a number of potential estimates (\eg by retrieving the $k$ most similar shapes in the training set~\cite{Shi25cvpr-CRISP}), in which case the probability $p\!\left(\{\MS_i\}_{i=1}^{N} \mid \MI, \MD, \{\MM_i\}_{i=1}^{N}\right)$ becomes a uniform distribution over discrete options. For simplicity, in the rest of this section we assume the network provides a single shape estimate and focus on pose estimation.
% $p\!\left(\MS_{[N]} \mid \MI, \MD, \MM_{[N]}\right)$ is a relatively harder problem compared with $p\!\left(\MT_{[N]} \mid \MI, \MD, \MM_{[N]}, \MS_{[N]}\right)$ \cite{chen25arxiv-sam}. Therefore, we can leverage any neural network to estimate the shape, \ie  model-agnostic to shape estimator. In Section \ref{sec:experiments}, we tested two shape estimators, a closed-set but relatively fast shape estimator and an open-set but relatively slow shape estimator. In the following sections, we will discuss the physics-grouned pose estimator of Picasso (Section \ref{sec:physics_pose_estimator}) and physics metrics to benchmark the performance of shape and pose estimation (Section \ref{sec:physics_metric}).

\myParagraph{Pose Likelihood} In contrast to the shape, we will expand the pose likelihood $p\!\left(\{\MT_i\}_{i=1}^{N} \mid \{\MS_i\}_{i=1}^{N}, \MI, \MD, \{\MM_i\}_{i=1}^{N}\right)$ analytically. Under the standard assumption of independent measurement noise, the probability factors as:
% \vspace{-1mm}
\begin{equation}
\label{eq:poseLike}
\begin{aligned}
p\!\left(\{\MT_i\}_{i=1}^{N} \mid \{\MS_i\}_{i=1}^{N}, \MI, \MD, \{\MM_i\}_{i=1}^{N}\right)
 = \\
 \prod_{i=1}^{N} p\!\left(\MT_i \mid \MS_i, \MI, \MD, \MM_i\right)
\end{aligned}
\end{equation}
% \vspace{-1mm}
where we also used the fact that an object's pose is independent from the masks and shapes of other objects.
Note that this does not mean the poses can be estimated independently; the resulting estimates are still coupled by the constraints in~\eqref{eq:registration}.

Now $p\!\left(\MT_i \mid \MS_i, \MI, \MD, \MM_i\right)$ is simply the likelihood of an object pose given its shape and measurements. Using RGB-D measurements, this can be framed as a point cloud registration problem.
In particular, if we denote as $\calA_i=\{\va_{i,j}\}_{j=1}^{M_{\calA_i}}$ the set of 3D points measured by the RGB-D camera which lie in the mask $\MM_i$ of object $i$, we can write the likelihood of the measurements as:
\begin{equation}
\label{eq:expFamily}
p\!\left(\MT_i \mid \MS_i, \MI, \MD, \MM_i \right) \propto  \exp\left(-\sum_{j=1}^{M_{\calA_i}} d\!\left(s_i \MR_i \va_{i,j}+\vt_i, \MS_i\right)\right)
\end{equation}
where the distribution is chosen to be in the exponential family (most commonly a Gaussian, if the noise is assumed to be additive and normally distributed), and depends on the distance $d\!\left(s_i \MR_i \va_{i,j}+\vt_i, \MS_i\right)$ of the measured point $\va_{i,j}$ (transformed to the object frame via $s_i, \MR_i, \vt_i$) to the surface of the object (as described by its shape $\MS_i$).

Substituting~\eqref{eq:expFamily} back into~\eqref{eq:poseLike} and observing that maximizing the likelihood is the same as minimizing the negative log-likelihood leads to:   
\begin{equation}
\begin{aligned}
\label{eq:NLL}
&\underset{\{\MT_i\}_{i=1}^{N} \in \calF}{\arg\max}\;
 \prod_{i=1}^{N} p\!\left(\{\MT_i\}_{i=1}^{N} \mid \{\MS_i\}_{i=1}^{N}, \MI, \MD, \{\MM_i\}_{i=1}^{N}\right) \\
=& \underset{\{\MT_i\}_{i=1}^{N} \in \calF}{\arg\min}\;
-\log  \prod_{i=1}^{N} p\!\left(\{\MT_i\}_{i=1}^{N} \mid \{\MS_i\}_{i=1}^{N}, \MI, \MD, \{\MM_i\}_{i=1}^{N}\right) \\
=&
\underset{\{\MT_i\}_{i=1}^{N} \in \calF}{\arg\min}\;
\sum_{i=1}^{N} \sum_{j=1}^{M_{\calA_i}} d\!\left(s_i \MR_i \va_{i,j}+\vt_i, \MS_i\right). &
\end{aligned}
\end{equation}

% \LC{get back sampling points on shape - currently we do not have defined the object point cloud}
While the expression of the distance in~\eqref{eq:NLL} depends on the shape representation (\eg SDF, mesh), 
a common approach is to sample a set of points $\calB_i=\{\vb_{i,k}\}_{k=1}^{M_{\calB_i}}$ from the object's surface 
and then use a point-to-point distance as metric, leading to the well-known Chamfer distance~\cite{Shi25cvpr-CRISP}:
% In the common case in which correspondences are not known, the problem becomes:
% \vspace{-1mm}
\begin{equation}
\begin{aligned}
d\!\left(s_i \MR_i \va_{i,j}+\vt_i, \MS_i\right) \doteq d\!\left(s_i \MR_i \va_{i,j}+\vt_i, \calB_i\right)\\
= \textstyle\sum_{j=1}^{M_{\calA_i}} \displaystyle\min_{1\leq k\leq M_{\calB_i}} \| s_i \MR_i \va_{i,j}+\vt_i - \vb_{i,k} \|^2 
%\nonumber
\label{eq:chamfer}
\end{aligned}
\end{equation}
% % \vspace{-1mm}
where, for each measured point $\va_{i,j}$, the objective computes the distance to the closest point in the object point cloud $\calB_i$. 
%Below we denote the Chamfer distance as $d\!\left(s_i \MR_i \va_{i,j}+\vt_i, \calB_i\right)$.
% This is typically referred to as the Chamfer distance~\cite{Shi25cvpr-CRISP}. 
%The full proof of the optimization objective is in Appendix \ref{app:registrationproof}.
% \vspace{-1mm}
% \[
% \min_{\MT_i \in \SIM(3)}\quad
% \sum_{j=1}^{M_{A,i}}
% \rho_{\text{GM}}\!\Big(
% d\!\left(s_i\,\MR_i\,\va_{i,j}+\vt_i,\;\calB_i\right)
% \Big),
% \]
% where $\rho_{\text{GM}}(d)=\frac{d^2}{d^2+\delta^2},\delta = 0.05\,\mathrm{m}$.

% If the correspondences between points in $\calA_i$ and $\calB_i$ were known, and assuming the (squared) Euclidean distance as point-to-point distance, problem~\eqref{eq:registration2} with $N=1$, can be solved in closed form using Horn's~\cite{Horn87josa} or Arun's method~\cite{Arun87pami}. 

\begin{remark}[The Single-Object Case]
\label{rmk:singleObject}
When $N=1$, \eqref{eq:NLL} reduces to standard point-cloud registration: estimate the transform aligning two point clouds.
% Problem~\eqref{eq:registration3} is a bilevel (nested) optimization problem over elements of the nonconvex $\SIM(3)$ group and is known to be hard to solve even in the single-object case~\cite{Yang20tro-teaser}.
% \begin{remark}[Hardness of Problem~\eqref{eq:registration3}]
% Problem~\eqref{eq:registration3} is a bilevel (nested) optimization problem over elements of the nonconvex $\SIM(3)$ group. The inner-level correspondence assignment depends on the transformation variables. An exhaustive search over correspondence assignments has a combinatorial space of size ${M_A}^{M_B}$, making the problem NP-hard.
% \end{remark}
A common solution strategy is to alternate between finding the correspondences between points in the two point clouds and computing the best transformation given the correspondences. 
% This leverages the fact that, given fixed correspondences, transformation estimation can be recovered in closed form (\cite{Horn87josa, Arun87pami, Li07iccv-3DRegistration}). 
For instance, the celebrated Iterative Closest Point (ICP) follows this paradigm. However, this line of approaches is prone to finding local minima. Other correspondence-free paradigms include optimization-based and sampling-based methods. Optimization-based methods phrase the problem as a mixed-integer program (MIP) and either solve it using MIP solvers~\cite{Izatt17isrr-MIPregistration}, Branch-and-Bound (BnB)~\cite{Breuel03cviu-BnBimplementation, Hartley09ijcv-globalRotationRegistration, Parra14cvpr-fastRotationRegistration, Yang16pami-goicp, gothoskar23arxiv-bayes3d}, or convex relaxations~\cite{Yang20tro-teaser}. While these are effective with strong correspondence priors settings~\cite{Yang20tro-teaser,Lim25icra-kissmatcher}, they fail in the correspondence-free case~\cite{Yang20tro-teaser}. 
Alternatively, sampling-based methods sample potential pose hypothesis until one that achieves a sufficiently small objective is found~\cite{Yang16pami-goicp}.  
% leverage the fact that the objective of~\eqref{eq:NLL} is easy to evaluate and hence sample potential pose hypothesis until one that achieves a sufficiently small objective is found~\cite{Yang16pami-goicp}.  
These approaches have a close relationship with MCMC~\cite{Gothoskar21arxiv-3dp3} and render-and-compare approaches~\cite{Labbe22corl-megapose, Wen24cvpr-FoundationPose}, which render  $\SE(3)$ transformation hypotheses and compare the rendered objects at those poses against the observed RGB or depth maps.  
\end{remark}

\begin{takeawaybox}
  In the single object case, sampling is an effective means to tackle problem~\eqref{eq:NLL} since (i) 
it only requires evaluating the highly nonconvex objective in~\eqref{eq:NLL}, (ii) we can sample from a relatively low-dimensional space ($\SIM(3)$), and (iii) evaluating each sample is trivially parallelizable.
%is a good fit when the objective is highly nonconvex but the feasible space is relatively low-dimensional.
\end{takeawaybox}

% \begin{remark}
% Sampling-based approaches have close relationship with render-and-compare approaches \cite{Labbe22corl-megapose, Wen24cvpr-FoundationPose} in 6-D pose estimation. They render a bunch of $\SE(3)$ 6-D transformation hypotheses and compare to the observed RGB and depth maps. Followed by a neural network refiner, they achieve fairly good pose estimation. Arguably, the insight of this type of works is that they leverage the prior from neural network to reduce the large search dimension and still be robust to outliers.
% \end{remark}

% \subsection{Haar Measures}

% \subsection{Physics-Grounded Multi-Object Pose Estimator}
%%%%%%%%%%%%%%%%%%%%%%%%%%%%%%%%%%%%%%%%%%%%%%%%%%%%%%%%%%%%%%%%%%%%%%
%%%%%%%%%%%%%%%%%%%%%%%%%%%%%%%%%%%%%%%%%%%%%%%%%%%%%%%%%%%%%%%%%%%%%%
\subsection{Physics Constraints}
\label{sec:constraints}

In this section, we derive the physics-based constraints  used in eq.~\eqref{eq:registration}. 
% Modeling \emph{dynamic} contact is challenging for several reasons \cite{le24tro-contact}. 
% First, rigid-body contact dynamics form a highly hybrid system: collisions induce discontinuous state changes, and the active contact set and modes (\eg  separation, sticking, sliding) can be difficult to detect reliably. 
% As a result, naive continuous-time integration is often unstable or inaccurate. 
% Second, even if all possible contact modes could be enumerated, solving for contact forces and mode transitions efficiently remains computationally demanding. 
% For this reason, most modern simulators adopt time-stepping approximations that enforce contact constraints at the discrete-time level.
We focus on \emph{static} scenes and restrict ourselves to configuration-space (holonomic) constraints, namely non-penetration and contact proximity. We assume the objects to lay on a planar surface (the \emph{environment}, \eg a table or ground floor), whose parameters we estimate beforehand (using a standard 3-point RANSAC from the depth measurements~\cite{Fischler81}). 
% This assumption comes without loss of generality in indoor environments, where we can assume  
% Extending our formulation to dynamic contact will be explored in future work. 
% Importantly, the non-penetration and contact constraints considered here admit simple closed-form expressions.

\begin{figure}[tb]

\centering
  \includegraphics[height=3.5cm,keepaspectratio]{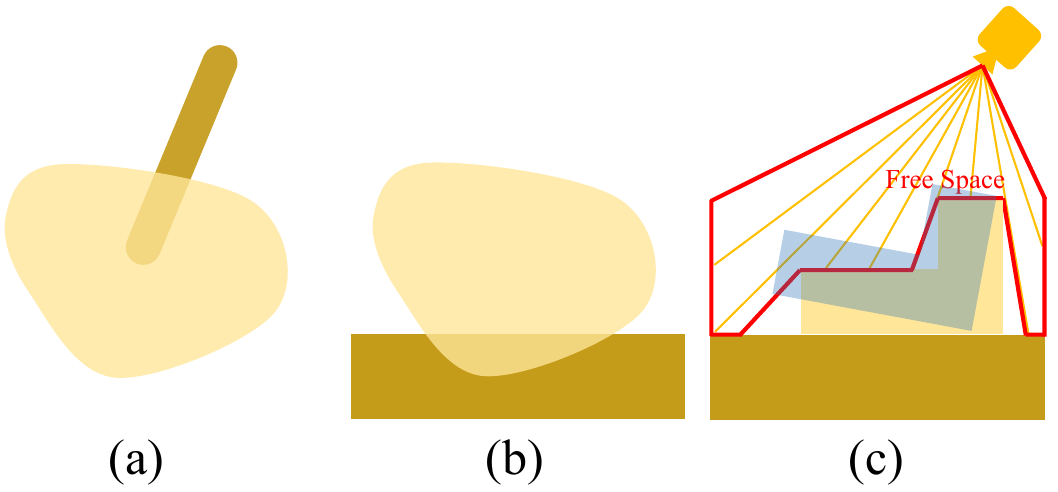}
\caption{(a) Inter-object penetration. (b) Object-environment penetration. (c) Object-free-space penetration: The estimated object (blue) penetrates the free space (red). Free space from an RGB-D image is defined as the empty volume along each camera ray up to the observed depth surface.
\label{fig:penetration}
\vspace{-4mm}}
\end{figure}

Let $\calB_i=\{\vb_{i,k}\}_{k=1}^{M_{\calB_i}}\subset\mathbb{R}^3$ denote vertices (or sampled surface points) 
in the local frame of object $i$ and recall that $\MT_i\in\SIM(3)$ maps points in the camera frame to the object frame.
Moreover, let $\Phi_i:\mathbb{R}^3\to\mathbb{R}$ be the signed distance field (SDF) of object $i$ in the local frame. 
% Each object pose $\MT_i\in\SIM(3)$ maps points in the camera frame to the object frame.
Similarly, let $\Phi_{\mathrm{env}}:\mathbb{R}^3\to\mathbb{R}$ denote the environmental SDF and $\Phi_{\mathrm{free}}:\mathbb{R}^3\to\mathbb{R}$ denote the free-space SDF (describing the geometry of the free space) respectively.
We enforce physical plausibility using four types of analytical constraints, including three types of penetration (visualized in Fig.~\ref{fig:penetration}) and one contact constraint: 

\paragraph{Inter-object non-penetration}
% We consider three types of penetration as shown in Fig.~\ref{fig:penetration}:
To prevent two objects $i$ and $j$ from intersecting, we require all transformed surface points of object $j$ to lie outside of object $i$, as measured by the corresponding SDF:
\begin{equation}
  \label{eq:inter_object_penetration}
\Phi_i\!\big(\MT_i\cdot\MT_j^{-1}\cdot\vb\big)\ge 0,
\ \forall\, i\neq j,\ \forall\, \vb\in\calB_j,
\end{equation}
where $\MT_i\cdot\MT_j^{-1}\cdot\vb$ denotes the 3D point $\vb$ (originally in the frame of object $j$), transformed to the frame of object $i$ using the similarity transforms $\MT_i\cdot\MT_j^{-1}$.

\paragraph{Object-environment non-penetration}
To prevent objects from penetrating the static environment (\eg  table or ground), we require every object point to remain outside the environment SDF denoted by $\Phi_{\mathrm{env}}$:
\vspace{-1mm}
\begin{equation}
  \label{eq:object_environment_penetration}
\Phi_{\mathrm{env}}\!\big(\MT_i^{-1}\cdot\vb\big)\ge 0,
\ \forall\, i,\ \forall\, \vb\in\calB_i.
\end{equation}

\paragraph{Object-free-space non-penetration}
To ensure object pose hypotheses are consistent with the observed free space (Fig.~\ref{fig:penetration}), we impose that object points should not enter regions known to be empty, represented by $\Phi_{\mathrm{free}}$:
\begin{equation}
  \label{eq:object_free_space_penetration}
\Phi_{\mathrm{free}}\!\big(\MT_i^{-1}\cdot\vb\big)\ge 0,
\ \forall\, i,\ \forall\, \vb\in\calB_i.
\end{equation}

\paragraph{Contact (no floating objects)}
Finally, we discourage floating objects by enforcing that each object must be in contact with at least one other object (or the environment) up to a small tolerance $\delta$:
% \vspace{-1mm}
\begin{equation}
  \label{eq:contact_constraint}
\forall\, i,\ \exists\, j\in[N]\!\setminus\!\{i\},\ \exists\, \vb\in\calB_i\ \text{s.t.}\
\big|\tilde{\Phi}_j(\MT_i^{-1}\cdot\vb)\big|\le \delta,
\end{equation}
% \vspace{-1mm}
where we define the world-frame SDF evaluator
% \vspace{-1mm}
\begin{equation}
\tilde{\Phi}_j(\bm{y}) := 
\begin{cases}
\Phi_{\mathrm{env}}(\bm{y}), & j=0,\\
\Phi_j\!\left(\MT_j\cdot\bm{y}\right), & j\in[N].
\end{cases}
\end{equation}
% \vspace{-1mm}
% We can finally combine the the objective we derived in eq.~\eqref{eq:NLL} and the constraints in eqs.~\eqref{eq:inter_object_penetration}-\eqref{eq:contact_constraint}.

% \begin{summarybox}
% General contact constraint can be hard to express, whereas deriving non-penetration and contact constraints for static scenes is comparatively straightforward. We can then check whether a candidate $\vtheta$ violates any constraint and reject it if so.
% \end{summarybox}

%\textbf{Multi-Object Pose Estimator. } 
%%%%%%%%%%%%%%%%%%%%%%%%%%%%%%%%%%%%%%%%%%%%%%%%%%%%%%%%%%%%%%%%%%%%%%
%%%%%%%%%%%%%%%%%%%%%%%%%%%%%%%%%%%%%%%%%%%%%%%%%%%%%%%%%%%%%%%%%%%%%%
\subsection{Physics-Constrained Pose Estimator}
\label{sec:optproblem}

We now rewrite the maximum likelihood pose estimator~\eqref{eq:registration} using the expression we derived for the objective in eqs.~\eqref{eq:NLL}-\eqref{eq:chamfer} and the constraints in eqs. (\ref{eq:inter_object_penetration}-\ref{eq:contact_constraint}).
This gives the following physics-constrained pose estimation problem:

\begin{plainbox}
\vspace{-10pt}
\begin{gather}
\hspace{-5mm}
\label{final_problem}
\begin{aligned}
\underset{\{\MT_i\}_{i=1}^{N} \in \SIM(3)^N}{\arg\min}\;\quad
& \sum_{i=1}^N \sum_{j=1}^{M_{A,i}}
d\!\left(s_i\MR_i \va_{i,j}+\vt_i,\calB_i\right) \\[2pt]
\text{s.t.}\quad
& \Phi_i\!\big(\MT_i\cdot\MT_j^{-1}\cdot\vb\big) \ge 0, \forall\, i\neq j,\ \forall\, \vb\in\calB_j, \\[2pt]
& \Phi_{\mathrm{env}}\!\big(\MT_i^{-1}\cdot\vb\big) \ge 0, \forall\, i,\ \forall\, \vb\in\calB_i, \\[2pt]
& \Phi_{\mathrm{free}}\!\big(\MT_i^{-1}\cdot\vb\big) \ge 0, \forall\, i,\ \forall\, \vb\in\calB_i, \\[2pt]
& \min_{j\in [N]\setminus\{i\}}\ \min_{\vb\in\calB_i}
\big|\tilde{\Phi}_j(\MT_i^{-1}\cdot\vb)\big| \le \delta, \forall\, i. \\[4pt]
% & \delta>0~\text{small}, \\
% & \tilde{\Phi}_j(\bm{y}) \coloneqq
% \begin{cases}
% \Phi_{\mathrm{env}}(\bm{y}), & j=0,\\
% \Phi_j\!\big(\MT_j\cdot\bm{y}\big), & j\in[N].
% \end{cases}
\end{aligned}
\raisetag{52pt}
\end{gather}
\vspace{-10pt}
\end{plainbox}
% where the existential statement is converted to the equivalent minimization problem for physics constraints.
 Problem~\eqref{final_problem} is highly nonconvex and optimizes $N$ similarity transformations---one per object. 
% In addition, the number of possible contacts for $N$ objects is $O(N^2)$, due to the existence of eq.~\eqref{eq:inter_object_penetration} and eq.~\eqref{eq:contact_constraint}, the number of constraints is also $O(N^2)$.
We will solve~\eqref{final_problem} using rejection sampling: that is, we sample a configuration of objects and enforce physical plausibility by simply rejecting any proposal that violates the constraints in~\eqref{final_problem}.
We check inter-object, object-environment, and contact constraints using the library~\cite{libigl}. 
For object-free-space constraint, we render the object as a depth map and directly compare it with observed depth map. 
 %In other words, we convert the constraint-satisfaction problem into a rejection-sampling procedure.
 Then, among the samples that satisfy the constraints, we select the one achieving the smallest objective.
% This approach leverages the fact that while~\ref{final_problem} is hard to solve globally, its objective and constraints are 
% relatively fast to evaluate.

A key remaining challenge is to reduce the dimensionality of the sampling space. 
Contrary to the single-object case discussed in Remark~\ref{rmk:singleObject}, naively solving Problem~\eqref{final_problem} via 
sampling would require high-dimensional samples, with a dimension that grows with the number $N$ of objects in the scene.
To address this, we leverage the notion of \emph{contact scene graph} 
to sequentially estimate each object's similarity transform via sampling.%, rather than sampling from a $7N$-dimensional space.
%and simplify problem~\eqref{final_problem} to a more tractable form in Section~\ref{sec:simplify_form}.

%%%%%%%%%%%%%%%%%%%%%%%%%%%%%%%%%%%%%%%%%%%%%%%%%%%%%%%%%%%%%%%%%%%%%%
%%%%%%%%%%%%%%%%%%%%%%%%%%%%%%%%%%%%%%%%%%%%%%%%%%%%%%%%%%%%%%%%%%%%%%
\subsection{Regaining Tractability via Contact Scene Graph}
\label{sec:simplify_form}

We start by recalling the classical Contact Scene Graph.
\begin{definition}[Contact Scene Graph (CSG)~\cite{hahn88scg-realistic}]
A \emph{contact scene graph} (CSG) is a tuple $\mathcal{SG} = (\calG, \vtheta)$ where $\calG = (\calV, \calE)$ is an undirected graph and $\vtheta$ are parameters associated with the nodes of the graph.
% We model the contact state of a scene as a \emph{scene graph} $\mathbf{\mathcal{G}}$, which is a tuple $\mathcal{SG} = (\calG, \vtheta)$ where $\calG = (\calV, \calE)$ is an undirected graph and $\vtheta$ are parameters associated to the nodes of the graph.
% $\MS_1,\ldots,\MS_N,\;\vt_1,\ldots,\vt_N$. 
\end{definition}

In a CSG, the vertices $\calV \coloneqq \{v_0, v_1, \ldots, v_N\}$ represent $N + 1$ rigid bodies ($v_0$ typically represents the environment, \eg a table for table-top scenes). Each vertex $v$ is associated with a parameter $\vtheta_v$ which consists of object pose and shape information. An edge $(u, v) \in \calE$ indicates that rigid body $u$ is in physical contact with rigid body $v$. That is, there exist points $\bm{p}_u, \bm{p}_v \in \mathbb{R}^3$ on the surface of rigid bodies $u$ and $v$ respectively such that $\bm{p}_u = \bm{p}_v$.  The CSG is undirected because if $u$ is in contact with $v$, then $v$ is in contact with $u$; however, it is not necessarily acyclic (\eg the set of Jenga blocks in the middle of Fig.~\ref{fig:dataset} form a ring of $n$ nodes). 

% Inference on CSG is defined as inferring parameter $\vtheta_v$ for $v \in \calV$ given sensor data, \eg  RGB and depth. 
% \begin{definition}[Inference on CSG]
% \label{inference}
% \end{definition}
Our goal, as described in the previous sections, is to estimate $\vtheta_v := (\MT_v, \MS_v)$. 
One could think about a CSG as a graphical model encoding relations between the variables of interest. 
However, 
% However, there are two difficulties for inferring $\vtheta$. First of all, extraction of $\calG$ out of raw sensor data is hard. One may consider pure geometry-based method like adjacency in depth point cloud but it loses human-level understanding of the environments \cite{schmid2024khronos}. Last but not the least, 
inference on an undirected cyclic graph is computationally hard~\cite{Cooper90ai}; on the other hand, inference over a directed acyclic graph (DAG) can be solved efficiently. We leverage this insight below: we first infer the CSG from an RGB image and then approximate it as a DAG; finally, we perform fast inference over the DAG using sampling.
%requires long mixing time. For example, random walk on the cycle of $N$ nodes requires mixing time at the order of $O(N^2)$ \cite{levin17AMS-markov}.

\textbf{Inferring and Approximating the Topology of the CSG.} 
%We mentioned the challenge of extracting $\bm{G}$ of the scene. 
The topology of the CSG is dictated by which objects are in contact. While a naive approach to infer contact would be to look at the 3D point clouds associated to each object and check if their distance is smaller than a threshold, we found this approach to be quite sensitive to noise and occlusions. For example, two separated objects can be mistakenly inferred as touching due to depth noise/outliers near edges, while true contacts (\eg object resting on a surface) can be missed entirely because the contact patch is occluded and thus absent from the point cloud.
On the other hand, we find that modern Vision-Language Models (VLMs) empirically give accurate answers to queries about contact. Therefore, we use a VLM to infer the topology of the CSG (see Appendix \ref{app:vlm} for details).
%, \eg  Gemini performed extraordinarily good at the semantic-level scene understanding. As for the long mixing time on CSG, we first acyclify the graph to construct Bayes Net. The acyclification process obeys the rule that environment is always treated as root. 
% TODO: worth describing further the implications of the  directed graph

\begin{figure}[t]
\centering
  \includegraphics[width=\columnwidth]{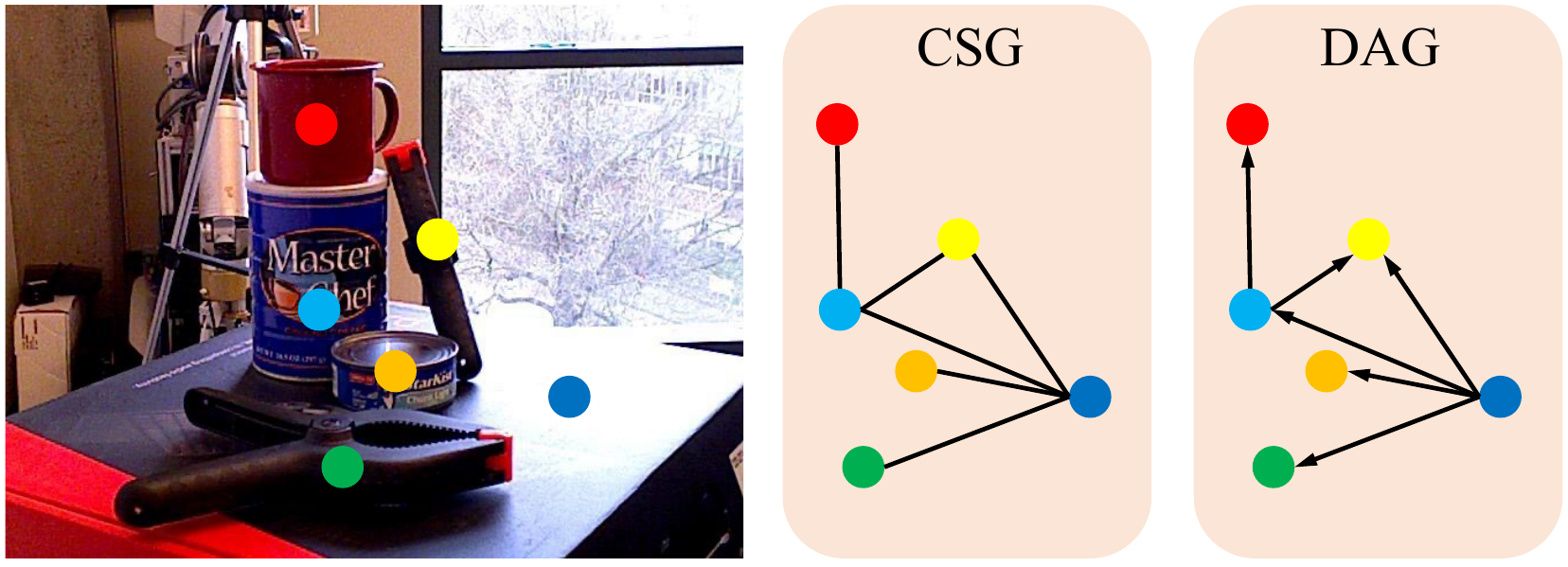}
\caption{Sample image and corresponding contact scene graph (CSG) with a directed acyclic graph (DAG) approximation.
\label{fig:contact}
\vspace{-6mm}}
\end{figure}

After inferring the full CSG, we approximate it with a DAG. We construct the DAG using breadth-first search and choose the environment node as the root. As we traverse the graph, we orient each edge in the order of traversal, yielding a directed graph. 
This structure enables a \emph{greedy} rollout: we estimate one object at a time (starting from the root) and treat 
the parents as fixed when 
inferring descendants. The number of inference calls on the DAG is exactly $N$ rather than $O(N)$ for inference on a full CSG.
%As a result, the number of inference calls is exactly $N$ rather than $O(N)$ for inference on undirected tree.
We emphasize that this is an approximation: greedy rollout 
is not guaranteed to recover the \emph{global} MLE assignment in~\eqref{final_problem}; however, 
we observe it achieves excellent accuracy in practice.

%Since every edge follows a single linear order, the directed graph is guaranteed to be acyclic.
Specifically, let $\calP(i) \subseteq \calV$ be the set of parents of node $i$.
After approximating the graph with a DAG, problem~\eqref{final_problem} decouples into $N$ subproblems, such that each subproblem is independent given the estimates of their parent nodes:
\begin{plainbox}
  \vspace{-10pt}
\begin{gather}
\label{decoupled_form}
\begin{aligned}
&\min_{\MT_i\in \SIM(3)}\quad
 \sum_{j=1}^{M_{A,i}}
d\!\left(s_i\MR_i \va_{i,j}+\vt_i,\calB_i\right) \\[2pt]
\text{s.t.}\quad
& \Phi_i\!\big(\MT_i\cdot\MT_j^{-1}\cdot\vb\big) \ge 0, \forall\, i\neq j, \ \forall\, \vb\in\calB_j, j \in \calP(i), \\[2pt]
& \Phi_{\mathrm{env}}\!\big(\MT_i^{-1}\cdot\vb\big) \ge 0, \forall\, i,\ \forall\, \vb\in\calB_i, \\[2pt]
& \Phi_{\mathrm{free}}\!\big(\MT_i^{-1}\cdot\vb\big) \ge 0, \forall\, i,\ \forall\, \vb\in\calB_i, \\[2pt]
& \min_{j\in \calP(i)\setminus\{i\}}\ \min_{\vb\in\calB_i}
\big|\tilde{\Phi}_j(\MT_i^{-1}\cdot\vb)\big| \le \delta, \forall\, i. \\[4pt]
% & \delta>0~\text{small}, \\
% & \tilde{\Phi}_j(\bm{y}) \coloneqq
% \begin{cases}
% \Phi_{\mathrm{env}}(\bm{y}), & j=0,\\
% \Phi_j\!\big(\MT_j\cdot\bm{y}\big), & j\in\calP(i).
% \end{cases}
\end{aligned}
\raisetag{55pt}
\end{gather}
  \vspace{-10pt}
\end{plainbox}

Therefore, we start by solving for the root of the tree in isolation. Then, following~\eqref{decoupled_form}, we solve for the immediate children of the root node in isolation, and so on. This is closely related to shock propagation in graphics simulation \cite{guendelman03tog-nonconvex}. 
Each subproblem in \eqref{decoupled_form} resembles the single-object registration problem in Remark~\ref{rmk:singleObject}, only with additional constraints. This makes it a natural fit for sampling-based optimization. 
We solve each subproblem via a coarse-to-fine sampling scheme over $\SIM(3)$, as discussed in the next section. % This suggests a simple inference algorithm. 
% We can traverse the nodes from root to leaves to solve the corresponding Problem~\eqref{decoupled_form} for each object $i$ given its parents. 

\subsection{Implementation of Coarse-to-Fine Rejection Sampling}
\label{sec:rejSampling}

We focus on the case where we already have a pose and shape estimator. 
Our approach takes as input the pose and shape estimate from an existing network 
(\eg we use SAM3D~\cite{chen25arxiv-sam} and CRISP~\cite{Shi25cvpr-CRISP} in our tests) 
and our sampling scheme corrects that estimate to make it more accurate and physically plausible.

\textbf{Initialization.} We initialize the object translation as the mean of the depth point clouds (restricted to the object mask), after removing outliers via 5\% thresholding. We initialize the object scale by comparing the mean depth of the masked point cloud to the mean depth of the rendered point cloud from the network's (\eg SAM3D) estimate.

\textbf{Sampler.} We solve each decoupled subproblem \eqref{decoupled_form} via a hierarchical sampler over $\SIM(3)$ by separating scale from $\SE(3)$ pose. 
We first perform a coarse-to-fine search over scale: starting from a broad set of scale hypotheses, 
we progressively refine around the best-performing candidates. For each scale hypothesis, 
we then run an $\SE(3)$ registration subroutine that samples translation and rotation in a coarse-to-fine manner. This coarse-to-fine strategy is also used in \cite{gothoskar23arxiv-bayes3d} and resembles BnB. 
%However, it does not come with a formal bound guarantee; analyzing the loss landscape to derive such guarantees is an interesting direction for future work.
Specifically, at each refinement level, we retain the top candidates according to the loss at the previous level and 
resample locally around them to increase resolution. 
We refer the reader to Appendix~\ref{app:samplingDetails} for further details. % and parameter choice.

\begin{figure}[t]
\centering
  \includegraphics[width=\columnwidth]{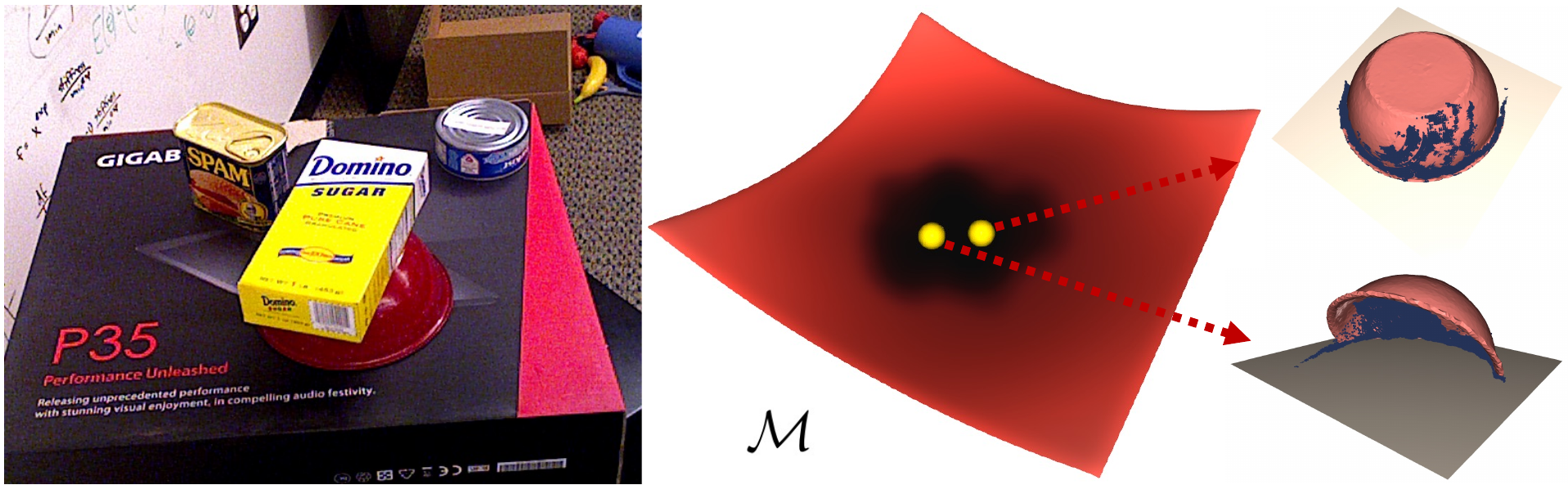}
\caption{Conceptual illustration of the loss landscape on the $\mathcal{M} = \SE(3)$ pose manifold for the red bowl shown in the left image. The landscape exhibits a region of similar loss values with two global minima (yellow balls) arising from the ambiguity of partial point cloud observations (dark blue). However, only one of these minima satisfies physics constraints and represents the correct pose. Physics constraints prune the feasible set to identify the physically valid solution.
%The shown surface is not the true visualization of the 
%$\SE(3)$ manifold; it is just an illustration.
\label{fig:loss_landscape}\vspace{-3mm}
}
\end{figure}

\textbf{Physics Constraints.} We use libigl~\cite{libigl} for object-object penetration and contact checking. For object-free-space penetration, we use the voxel-based depth renderer implemented in \cite{Paszke19neurips-pytorch}. 
We apply rejection sampling to the top $B=16$ candidates ranked by Chamfer loss. If all $B$ violate constraints, we return the best-scoring pose. A visualization of how physics constraints can help pose estimation is shown in Fig.~\ref{fig:loss_landscape}.

\textbf{Robust Loss.}
In practice, we augment the objective in~\eqref{decoupled_form} with a robust loss to mitigate the effect of segmentation artifacts, shape prediction errors, and depth sensor noise. In particular, we use the Geman-McClure robust loss~\cite{Barron19cvpr-adaptRobustLoss} 
to robustify the Chamfer distance, namely $\rho_{\text{GM}}\!\big(
d\!\left(s_i\,\MR_i\,\va_{i,j}+\vt_i,\;\calB_i\right) \big)$, where $\rho_{\text{GM}}(d)=d^2 / (d^2+\delta^2),\delta = 0.05\,\mathrm{m}$.

\textbf{System.} Overall, this results in a unified, sampling-based solver for physics-constrained $\SIM(3)$ registration, which we name \textit{Picasso}. 
Picasso is a pose corrector, as it leverages rough poses for object scale initialization.
%To the best of our knowledge, Picasso is the first physics-constrained, sampling-based pipeline for holistic scene reconstruction, and the first sampling-based solver for $\SIM(3)$
% registration. 
Picasso enables us to estimate each object in a fraction of a second using GPU-optimized parallel sampling (Section \ref{sec:close_set}), 
%it enables \textbf{real-time} single object physics-constrained $\SE(3)$ registration as shown in Section \ref{sec:close_set} 
and outperforms state of the art estimators (Section~\ref{sec:experiments}).
\begin{takeawaybox}
Picasso (i) decouples holistic scene understanding into a sequential estimation of object poses, while accounting
for physical constraints among objects, and (ii) enables fast inference using GPU-parallelized coarse-to-fine rejection sampling. 
% (1) Picasso achieves fast and holistic scene understanding with physics-constrained sampling. \\
% (2) Fast registration is enabled by GPU-parallelized Bayesian inference. \\
\end{takeawaybox}

%% file: sections/dataset.tex
%!TEX root = ../main.tex

\section{Picasso Dataset and Physics Metric}
\label{sec:dataset}
We now introduce the \emph{Picasso dataset}, a new object-centric dataset focused on contact-rich real-world scenes. We also define a \emph{scene plausibility score} (SPS), which is designed to measure the physical plausibility of a scene reconstruction.

\subsection{The Picasso Dataset}

\myParagraph{Motivation} Object-centric datasets provide ground-truth annotations for tasks such as object segmentation, pose, and shape. Large-scale object shape repositories are now widely available~\cite{Chang15arxiv-shapenet,deitke23neurips-objaverse,reizenstein21iccv-common,wu23cvpr-omniobject3d,downs22icra-google}, and several datasets offer pose and shape annotations, including NOCS and YCB-Video~\cite{Wang19-normalizedCoordinate,Xiang17rss-posecnn}. However, most existing datasets do not model physical interactions and contacts explicitly. The closest related effort is Vysics~\cite{bianchini25arxiv-vysics}, which focuses on contact-rich tabletop trajectories. Our Picasso dataset provides not only object shapes but also per-object 6D pose annotations, and targets {RGB-D single-image estimation rather than object tracking.}
% LS: this isn't novel as is. Does Picasso also provide contacts / model physical interactions in some way?

\myParagraph{Dataset Annotation}
We capture RGB-D data using an iPad's built-in LiDAR and camera sensors at $1920\!\times\!1440$ resolution and 60\,Hz. Object masks are produced automatically using SAM2~\cite{Ravi24arxiv-sam2} and then verified and corrected by human annotators. Subsequently, the object models are scanned and refined in Blender~\cite{Blender18-blender}. To obtain 3D poses, we solve PnP by labeling 2D keypoints in the images and establishing correspondences to keypoints on the scanned 3D object models. Finally, we refine the raw depth maps by rendering the object models under the annotated poses.

\myParagraph{Dataset Characteristics}
The Picasso dataset contains 10 contact-rich static scenes featuring 10 everyday objects (Fig.~\ref{fig:dataset}). For each scene, we select 10 images and annotate the pose and shape of every object. The number of objects per scene ranges from two to eight. The scenes cover diverse contact types, including point contacts (\eg mugs supported by a rack), line contacts (\eg a spatula resting on a pan), and flush contacts (\eg stacked Jenga blocks). 
%Our annotations may also be affected by several sources of error, including LiDAR depth noise, inaccuracies in the scanned object models, and residual errors from the PnP solver.

\subsection{Physics Metric}
\label{sec:physics_metric}

We define the \emph{Scene Plausibility Score (SPS)} to ascertain  scene-level physics plausibility. 
 Given a set of pose and shape estimates of the scene, we create a digital twin in 
 simulation\footnote{We used PyBullet in this paper but any other simulator can be used.} \cite{wu15nips-galileo} and run a short rollout of $T$ steps. For all scenes, we use $T=20$ with a simulation timestep of $1/240s$. Then we compute SPS:
\begin{equation}
S^{SPS}
:= \frac{1}{2}m \vv\tran \vv
+ \frac{1}{2}\M{\omega}\tran \M{J} \M{\omega}
\label{eq: sps}
\end{equation}
where $\vv$ and $\M{\omega}$ are the linear and angular velocity of the object. %'s center of mass. 
We set nominal mass $m=1$\,kg and inertia $\M{J}=\eye_3$ (identity matrix). The metric measures the kinetic energy of a system with this canonical mass and inertia. Due to potential inaccuracies in the simulator's physical modeling, we threshold translational and rotational kinetic energies to a maximum of 10 $kg \cdot m^2/s^2$ in all experiments.
% \begin{remark}
% We simulate only a very short time horizon because we found that, given the simulator’s imperfect physics engine, longer rollouts tend to accumulate noise and drift.
% \end{remark}
% \begin{remark}
% We do not include translation or rotation explicitly in the energy computation because. This is because over the short horizon of $1/12s$, the linear and angular velocities are first-order derivatives of translation and rotation.
% \end{remark}

%% file: sections/experiment.tex
%!TEX root = ../main.tex

\section{Experiments}
\label{sec:experiments} 

% We show that Picasso can be easily retrofitted on existing pose and shape estimators, including CRISP~\cite{Shi25cvpr-CRISP} (Section~\ref{sec:close_set}) and SAM3D~\cite{chen25arxiv-sam} (Section~\ref{sec:open_set}) and show it 
% achieves state of the art performance on the YCB-V and Picasso datasets, while better aligning with human 
% physical intuition (Section~\ref{sec:humans}).
We evaluate Picasso in a variety of real-world settings and retrofit it to CRISP~\cite{Shi25cvpr-CRISP}  and SAM3D~\cite{chen25arxiv-sam}.
Section~\ref{sec:close_set} shows that Picasso bridges the sim-to-real gap for CRISP, allowing a simulation-trained approach to outperform an approach trained on real data.
Section~\ref{sec:open_set} shows Picasso improves SAM3D performance on the YCB-V and Picasso datasets.
% Section~\ref{sec:humans} shows Picasso's reconstructions align with human physical intuition and validates our proposed SPS metric.
%  (Section~\ref{sec:open_set}).
% with several existing pose and shape estimators. In Section \ref{sec:close_set}, we show that Picasso brings CRISP~\cite{Shi25cvpr-CRISP} across the sim-to-real gap, outperforming CRISP trained on real images. We also show Picasso with SAM3D~\cite{chen25arxiv-sam} achieves state of the art performance on the YCB-V and Picasso datasets (Section~\ref{sec:open_set}), while better aligning with human physical intuition (Section~\ref{sec:humans}).

\subsection{Boosting CRISP's Performance Using Picasso}
\label{sec:close_set}

\textbf{Setup.} We train two models: CRISP-Real, which is trained on the real-world training images provided by the YCB-V dataset, and CRISP-Syn, which is trained on 4,200 synthetic images (200 per object) rendered with BlenderProc \cite{Denninger20rss-blenderproc} as provided by \cite{Shi25cvpr-CRISP}. We use the synthetic setup to assess Picasso's ability to
correct rough estimates caused by a large sim-to-real gap. In addition to the two CRISP variants, we compare against two baselines: (i) A physics-guided approach that refines CRISP by incorporating non-penetration losses, optimized via gradient descent (more details in Appendix \ref{app:physicsloss}), (ii) a Generalized ICP-based pose corrector, as in \cite{Koide24joss-smallgicp}. We report ADD-S~\cite{Xiang17rss-posecnn}, observable correctness (OC)~\cite{Shi25cvpr-CRISP}, and the proposed Scene Plausibility Score  (SPS). We also report
the Non-Penetration Score (NPS)~\cite{zhang22icra-non} with a small modification to exclude bad shape estimates (Appendix~\ref{app:nps}).

\textbf{Results.} As shown in Table \ref{tab:closed_set}, CRISP-Real outperforms \cite{Labbe20eccv-CosyPose, Wen23cvpr-bundlesdf, Liu22eccvw-gdrnppBOP} and CRISP-Syn, indicating a large sim-to-real gap. With Picasso, however, CRISP-Syn becomes the top-performing method, outperforming local pose refinement such as gradient descent and GICP \cite{Segal09rss-GeneralizedICP}, and even surpassing CRISP-Real to close the sim-to-real gap. 
Table \ref{tab:ycbv_phy_time} shows that Picasso also delivers the best results on physics metrics (NPS and SPS) and achieves fast performance with runtime under 1 second.

% ================= Table 2 =================
\begin{table}[t]
\centering
\setlength{\tabcolsep}{6pt}
\caption{Evaluation results on the ADD-S (m) and ADD-S (AUC) metrics for the YCB-V dataset.  \colorbox{green!60}{Best}, \colorbox{yellow!60}{Second-best}.
%CRISP-Syn+P: CRISP-Syn with Picasso corrector.
}
\label{tab:closed_set}
\begin{tabular}{lccccc}
\hline
 & \multicolumn{2}{c}{ADD-S $\downarrow$} & \multicolumn{3}{c}{ADD-S (AUC) $\uparrow$} \\
\hline
Method & Mean & Median & 1 cm & 2 cm & 3 cm \\
\hline
CosyPose \cite{Labbe20eccv-CosyPose}               & 0.010 & 0.007          & 0.30 & 0.56 & 0.68 \\
BundleSDF \cite{Wen23cvpr-bundlesdf}              & 0.014          & 0.012          & 0.14 & 0.37 & 0.55 \\
GDRNet++ \cite{Liu22eccvw-gdrnppBOP}               & 0.013          & 0.011          & 0.22 & 0.43 & 0.58 \\
CRISP-Real \cite{Shi25cvpr-CRISP}                  & {0.009}   & {0.004} & 0.44 & 0.58 & 0.75 \\
CRISP-Syn  \cite{Shi25cvpr-CRISP}                 & 0.015          & 0.010          & 0.21 & 0.41 & 0.55 \\
CRISP-Syn+GD          & 0.015   &  0.011  & 0.21 & 0.41 & 0.55 \\
CRISP-Syn+GICP          & 0.012          & 0.003   & 0.44 & 0.58 & 0.67 \\
CRISP-Syn+Picasso             & \second{0.008}   & \second{0.003}   & \second{0.52}   & \second{0.69} & \second{0.77} \\
CRISP-Real+Picasso                  & \best{0.008}   & \best{0.003} & \best{0.53} & \best{0.70} & \best{0.78} \\
\hline
\end{tabular}
\end{table}

% ================= Table 3 =================
\begin{table}[t]
\centering
\setlength{\tabcolsep}{6pt}
\caption{Evaluation results on the SPS, NPS (mm), observable correctness, and runtime on the YCB-V dataset. \colorbox{green!60}{Best}, \colorbox{yellow!60}{Second-best}.
%CRISP-Syn+P: CRISP-Syn with Picasso corrector.
}
\label{tab:ycbv_phy_time}
\begin{tabular}{lcccc}
\hline
Method  & {SPS $\downarrow$} & {NPS $\downarrow$} & {OC $\uparrow$} & {run time (s) $\downarrow$} \\
\hline
CRISP-Real & 9.05 & 4.57 & $36\%$ & /\\
CRISP-Syn & 9.36 & 6.38 & $15\%$ & /\\
CRISP-Syn+GICP          & 9.56              & 4.70           & $42\%$         & \best{0.03} \\
CRISP-Syn+Picasso               & \best{8.71} & \second{3.99} & \best{$55\%$} & 0.35      \\
CRISP-Real+Picasso               & \second{8.77} & \best{3.95} & \second{$55\%$} & \second{0.33}      \\
\hline
\end{tabular}
\end{table}

\subsection{Boosting SAM3D's Performance Using Picasso}
\label{sec:open_set}

We compare against SAM3D \cite{chen25arxiv-sam}, a foundation model that predicts object shape and 6D pose (up to scale) from monocular images. To estimate scale for SAM3D, we compute the ratio between the mean depth of the masked depth map and the mean rendered depth from SAM3D's predicted pose and shape. For Picasso, we take the SAM3D estimates and compute a similarity transform as discussed in Section~\ref{sec:method}.
%enable the per-object $\SIM(3)$ scale estimation module. 

\subsubsection{YCB-V Dataset}
We test on the YCB-V dataset from the BOP19 challenge consisting of 900 frames.

\textbf{Results.} We compare SAM3D (S) and SAM3D+Picasso (S+P) in Table \ref{tab:ycbvideo}. We also compare with SAM3D+Picasso {without physics constraints (S+P-w/o PC)} (\ie only sampling from $\SIM(3)$ to minimize the objective). Picasso significantly improves SAM3D in both a geometric metric (ADD-S) and physical metric (NPS and SPS). Picasso also outperforms the ablation without physics constraints, but the benefit is more modest because we only check physics plausibility on a small buffer of candidates ranked by Chamfer loss. 
%Future work will extend the plausibility check to a much larger set of candidates.

% We also visualize some qualitative results. 
% An interesting phenomenon is that geometrically accurate reconstructions may not lead to physically valid configuration? The answer is no. When we compare S+C against S+PC.

\begin{table}[t]
\centering
\caption{Comparison of methods across 12 YCB-Video trajectories. ADD-S (mm), SPS (physics loss), NPS (mm). S: SAM3D. S+P-w/o PC: SAM3D+Picasso w/o physics constraints. S+P: SAM3D+Picasso. \colorbox{green!60}{Best}.
}
\label{tab:ycbvideo}
\resizebox{\columnwidth}{!}{%
\begin{tabular}{l|ccc|ccc|ccc}
\toprule
& \multicolumn{3}{c|}{\textbf{ADD-S} $\downarrow$} & \multicolumn{3}{c|}{\textbf{SPS} $\downarrow$} & \multicolumn{3}{c}{\textbf{NPS} $\downarrow$} \\
\textbf{Trajectory} & S\cite{chen25arxiv-sam} & \makecell{S+P\\-w/o PC} & S+P & S\cite{chen25arxiv-sam} & \makecell{S+P\\-w/o PC} & S+P & S\cite{chen25arxiv-sam} & \makecell{S+P\\-w/o PC} & S+P\\
\midrule
traj\_48 & 16.97 & \cellcolor{green!60}8.98 & 9.06 & 15.32 & 5.38 & \cellcolor{green!60}5.16 & 10.13 & 2.82 & \cellcolor{green!60}2.76\\
traj\_49 & 29.53 & 26.74 & \cellcolor{green!60}26.59 & 4.21 & 2.32 & \cellcolor{green!60}2.14 & 7.94 & 5.03 & \cellcolor{green!60}4.72\\
traj\_50 & \cellcolor{green!60}11.77 & 19.91 & 20.48 & 1.60 & \cellcolor{green!60}0.86 & 1.14 & 5.65 & 3.86 & \cellcolor{green!60}3.53\\
traj\_51 & 13.25 & \cellcolor{green!60}7.40 & 7.84 & 8.51 & \cellcolor{green!60}2.87 & 3.01 & 7.31 & 2.37 & \cellcolor{green!60}2.13\\
traj\_52 & 17.78 & \cellcolor{green!60}6.28 & 6.76 & 6.16 & 0.12 & \cellcolor{green!60}0.12 & 16.53 & 3.20 & \cellcolor{green!60}2.91\\
traj\_53 & 16.07 & \cellcolor{green!60}5.17 & 5.46 & 6.50 & 0.04 & \cellcolor{green!60}0.04 & 11.51 & 3.23 & \cellcolor{green!60}2.98\\
traj\_54 & 22.94 & 10.22 & \cellcolor{green!60}10.19 & 10.77 & \cellcolor{green!60}4.53 & 4.69 & 10.70 & 2.51 & \cellcolor{green!60}2.47\\
traj\_55 & 68.73 & \cellcolor{green!60}9.53 & 10.23 & 3.60 & 0.63 & \cellcolor{green!60}0.63 & 11.11 & 2.16 & \cellcolor{green!60}1.95\\
traj\_56 & 16.74 & \cellcolor{green!60}9.21 & 9.45 & 4.42 & 1.48 & \cellcolor{green!60}1.19 & 9.01 & 2.77 & \cellcolor{green!60}2.67\\
traj\_57 & 19.48 & \cellcolor{green!60}6.88 & 6.95 & 2.08 & 0.48 & \cellcolor{green!60}0.39 & 5.12 & 1.94 & \cellcolor{green!60}1.69\\
traj\_58 & 13.06 & \cellcolor{green!60}7.11 & 7.31 & 3.43 & 3.18 & \cellcolor{green!60}2.73 & 10.27 & 4.16 & \cellcolor{green!60}3.86\\
traj\_59 & 13.81 & \cellcolor{green!60}6.11 & 6.46 & 3.54 & 1.03 & \cellcolor{green!60}1.02 & 6.26 & 1.96 & \cellcolor{green!60}1.84\\
\midrule
\textbf{Overall} & 21.68 & \cellcolor{green!60}10.29 & 10.56 & 5.84 & 1.91 & \cellcolor{green!60}1.85 & 9.29 & 3.00 & \cellcolor{green!60}2.79\\
\bottomrule
\end{tabular}%
}
\vspace{-3mm}
\end{table}

\subsubsection{Picasso Dataset} We further test on the Picasso dataset, which contains 100 frames. Specifically, we compare against: (1) SAM3D with RGB input (denoted as SAM3D),
(2) SAM3D with both RGB and pointmap input (denoted as
SAM3D-PM), and (3) SAM3D-PM combined with GICP. 
% Otherwise, we use the same experimental setup as in the YCB-V experiment.

\begin{figure}[t]
\centering
  \includegraphics[width=\columnwidth]{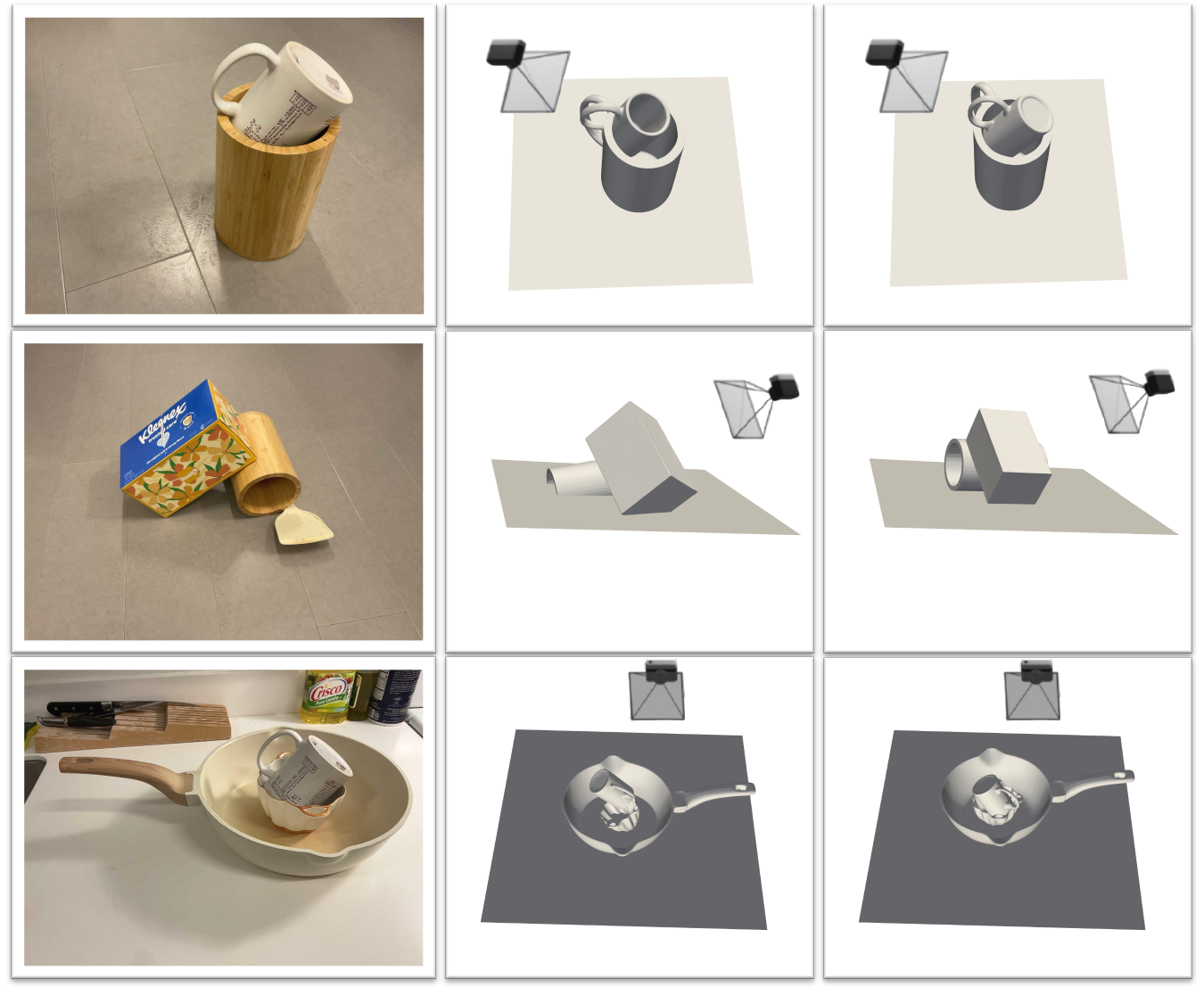}
\caption{Visualization of 3D scene reconstruction from the Picasso Dataset. \textbf{Left: }  Input RGB images. \textbf{Middle:} 
SAM3D reconstructions. \textbf{Right:} SAM3D+Picasso reconstructions. While SAM3D reconstructions have 
severe penetrations and floating objects, Picasso produces physically plausible configurations. 
\label{fig:picasso_vis} \vspace{-5mm}}
\end{figure}

\textbf{Results.} Fig.~\ref{fig:picasso_vis} shows qualitative 3D reconstructions. 
While SAM3D produces severe inter-object penetrations and counter-intuitive floating objects, 
Picasso yields physically valid results. As shown in Table~\ref{tab:picasso}, Picasso consistently improves
performance across all metrics under both the SAM3D and
SAM3D-PM settings by a large margin, and it outperforms
the registration method GICP on all but one metric.

\begin{table}[t]
\centering
\caption{Comparison of methods across 10 Picasso sequences. ADD-S (mm), SPS (physics loss), NPS (mm). SAM3D: SAM3D with RGB input, 
SAM3D-PM: SAM3D with both RGB and pointmap input. \colorbox{green!60}{Best}, \colorbox{yellow!60}{Second-best}.}
\label{tab:picasso}
\resizebox{\columnwidth}{!}{%
\begin{tabular}{lccccccc}
\hline
\multirow{2}{*}{Method} 
& \multicolumn{2}{c}{ADD-S (mm) $\downarrow$} 
& \multicolumn{3}{c}{ADD-S (AUC) $\uparrow$} 
& \multirow{2}{*}{SPS $\downarrow$} 
& \multirow{2}{*}{NPS $\downarrow$} \\
\cline{2-6}
& Mean & Median & 1 cm & 2 cm & 3 cm & & \\
\hline
SAM3D-PM  & 11.45 & 6.49 & 33.59 & 56.17 & 68.35 & 8.91 & 5.67 \\
+GICP     & \cellcolor{green!60}7.23  & 5.77 & 43.82 & 69.04 & 78.93 & 8.58 & 4.32 \\
+Picasso (Ours)  & 7.84  & \cellcolor{yellow!60}5.18 & \cellcolor{yellow!60}47.07 & \cellcolor{green!60}70.99 & \cellcolor{yellow!60}79.73 & \cellcolor{green!60}6.59 & \cellcolor{green!60}3.50 \\
\hline
SAM3D     & 11.71 & 7.52 & 30.60 & 53.61 & 66.53 & 8.50 & 5.23 \\
+Picasso (Ours) & \cellcolor{yellow!60}7.57  & \cellcolor{green!60}4.95 & \cellcolor{green!60}47.38 & \cellcolor{yellow!60}70.91 & \cellcolor{green!60}79.77 & \cellcolor{yellow!60}7.63 & \cellcolor{yellow!60}3.74 \\
\hline
\end{tabular}%
}
\end{table}

\subsection{User Study: Alignment with Physics Metrics}
\label{sec:humans}
% LS: this section is not particularly convincing. Much of the setup could probably fit in the appendix. The results are not conclusive and I'm not certain it adds too much to the paper.
% Also: the Public is basically 50/50, right? Hard to know what the percentages mean

Finally, we investigate the question: \emph{Do Picasso's reconstructions align with human physical intuition?}
Towards this goal, we perform a user study using Amazon Mechanical Turk, where users rank the plausibility of 3D reconstructions given videos of the corresponding digital twins. The results show that (i) human evaluation aligns well with our Scene Plausibility Score (SPS), and (ii) Picasso's reconstructions are ranked consistently better than SAM3D results in terms of physical plausibility. 
We refer the reader to Appendix~\ref{sec:humansAppendix} for details about the data collection and results.

%% file: sections/limitations_future_work.tex
%!TEX root = ../main.tex

\section{Limitations and Future Work}
\label{sec:limitationsApp}
We conclude by outlining the limitations of Picasso and highlighting several promising directions for future research:

\paragraph{Alternative Objectives}
Although we focus on point-cloud registration, physics-constrained sampling naturally extends to other highly nonconvex objectives. For instance, it can be combined with semantic-based or rendering-based losses~\cite{Labbe22corl-megapose, gothoskar23arxiv-bayes3d, zhou23iccv-3d, Wen24cvpr-FoundationPose}, or with latent-space objectives that compare observations and renderings through learned embeddings \cite{dupont22arxiv-data, bauer23arxiv-spatial}.

\paragraph{Accelerating the Solver}
We plan to investigate faster sampling-based solvers, for example by leveraging BnB and 3D Euclidean Distance Transforms for efficient scoring~\cite{Yang16pami-goicp}.

\paragraph{Improving Contact Scene Graph Inference}
Our current approach makes the contact graph acyclic and performs local MLE rollout. Future work will relax this approximation to support mutual interactions between upstream and downstream nodes, while preserving fast inference~\cite{guendelman03tog-nonconvex}.

\paragraph{Physics-Constrained Foundation Models}
We applied Picasso to improve inference-time performance of foundation models, such as SAM3D.
% Physics constraints may guide inference in other generative models ---for example, shape prediction models where candidates are sampled from a learned distribution~\cite{huang25cvpr-spar3d, chen25arxiv-sam}. 
Beyond inference, another direction is to use Picasso to generate constraint-satisfying data that could be used to augment training for foundation models. In addition, such physics constraints could be incorporated directly into the model training procedure, building on recent work in constrained diffusion modeling~\cite{fishman23arxiv-diffusion, nordenhog25arxiv-score, blanke25arxiv-strictly}. 

%% file: sections/conclusion.tex
%!TEX root = ../main.tex

\section{Conclusion}
\label{sec:conclusion}

We proposed an approach for holistic scene understanding, which accounts for sensor measurements (\eg RGB-D scans) and physical plausibility.
The proposed approach, Picasso, decouples scene reconstruction
into a sequential physics-informed estimation of object poses, and
 enables fast inference using GPU-parallelized
coarse-to-fine rejection sampling.
% In this paper, we posit that object pose and shape estimation requires reasoning holistically over the scene (instead of reasoning about each object in isolation) and accounting for object interactions and physical plausibility. % of the estimated scene.
% Towards this goal, our first contribution is \emph{Picasso}, 
% % In this project, we ask: \emph{Can we build a scene reconstruction system that not only achieves geometric accuracy, but also enforces physical plausibility?} To this end, we propose 
% a physics-constrained reconstruction pipeline that builds multi-object scene reconstructions by accounting for geometry, non-penetration, and physics of the scene.
% Picasso relies on a fast rejection sampling method that reasons over multi-object interactions in a tractable way by leveraging the fact only a subset of objects are in mutual contact in the scene. 
%, which helps breaking down the complexity of reasoning over mutual interactions.
%based on fast rejection sampling over a contact scene graph.
We also proposed the \emph{Picasso dataset}, a collection of 10 contact-rich real-world scenes with ground truth annotations, 
as well as a novel metric to quantify physical plausibility, which we open-sourced with our benchmark. 
Results on the Picasso and YCB-V dataset show that Picasso largely  outperforms the state of the art while providing reconstructions that are both physically plausible and more aligned with human intuition.
%  We refer the reader to Appendix~\ref{sec:limitationsApp} for further discussion about limitations and future work.

%% file: sections/appendix.tex
\appendix
\section{Appendix}

\subsection{VLM for CSG Generation}
\label{app:vlm}
\begin{figure}[t!]
\centering
  \includegraphics[width=\columnwidth]{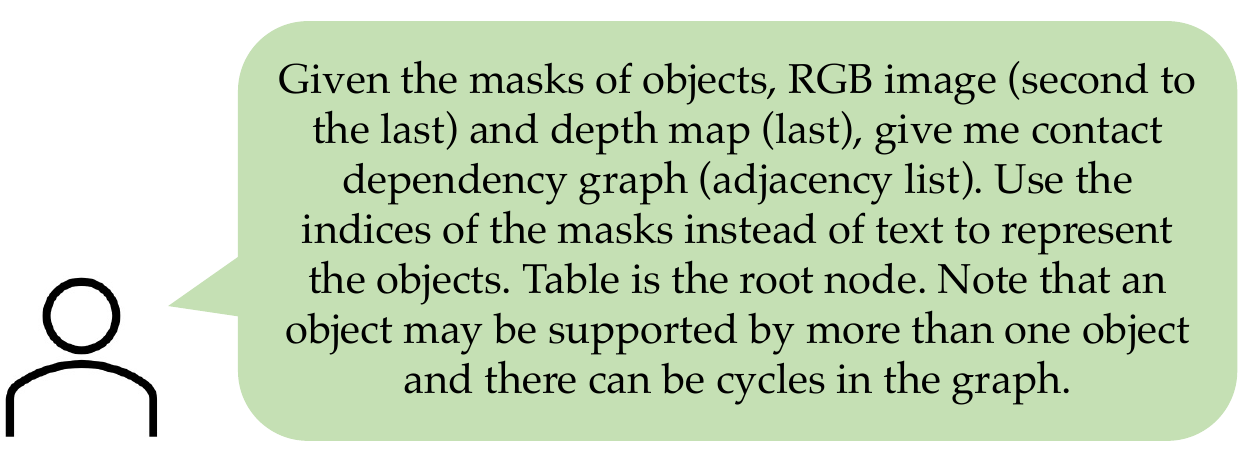}
\caption{VLM prompt for contact scene graph generation.}
\label{fig:vlm}
\end{figure}

In this section, we explain how we construct a contact scene graph (CSG) for an arbitrary scene using a vision-language model (VLM). Empirically, we find that VLMs can accurately estimate object contacts using only a single RGB image. We provide a VLM with object masks, the RGB image, and the depth map, and we query it for a contact graph between the object masks. 
We use the prompt given in Fig.~\ref{fig:vlm}. 
We use Google Gemini throughout all experiments. 

% We empirically find that VLMs accurately decide which objects are in contact from a single image view. 

\subsection{Coarse-to-Fine Sampling Details}
\label{app:samplingDetails}
\begin{table}[t]
\centering
\small
\setlength{\tabcolsep}{8pt}
\begin{tabular}{c c c c c}
\toprule
Level & Type & $\theta_{\max}$ & $\theta_{\text{step}}$ & \# rotations per seed \\
\midrule
1 & Global & -- & -- & 1024 \\
2 & Tangent Space & $30^\circ$ & $6^\circ$ & 515 \\
3 & Tangent Space  & $6^\circ$  & $2^\circ$ & 123 \\
4 & Tangent Space  & $3^\circ$  & $2^\circ$ & 19 \\
\bottomrule
\end{tabular}
\caption{Hierarchical rotation sampling schedule. Level 1 performs global sampling on $\mathrm{SO}(3)$; later levels refine locally around propagated seeds.}
\label{tab:rot-grid}
\end{table}

As we mentioned in Section~\ref{sec:rejSampling}, we first perform a coarse-to-fine 
search to estimate scale, rotation, and translation. 
For each scale hypothesis, we run an $\SE(3)$ registration 
subroutine that jointly samples translation and rotation, progressively 
refining and sampling around the best-performing candidates. 

To find scale, we use three levels spanning $[0.5, 1.1]$ with $K=5$ uniform samples per level, and refine each subsequent level around the best candidate.
For rotation, we adopt four levels. At the coarsest level, we first sample 1024 rotations uniformly on $\mathrm{SO}(3)$. At subsequent levels, we take seed rotations from the top candidates of the previous level and refine locally. Given a seed rotation $\MR_{\text{seed}}$, we sample local perturbations in the tangent space. That is,
$\MR_{\text{local}} = \MR_{\text{seed}} \cdot \exp(\boldsymbol{\omega}),$ where $\exp(\cdot)$ is the exponential map and $\boldsymbol{\omega}\in\mathbb{R}^3$ is a tangent space vector with elements $\omega_i \in [-\theta_{\max}, \theta_{\max}]$. Then we conduct a local grid search with step size $\theta_{\text{step}}$. The local search window $(\theta_{\max}, \theta_{\text{step}})$ decreases per level and gets finer as shown in Table~\ref{tab:rot-grid}.
Lastly, for translation we normalize the object to unit scale using the predicted shape and estimate translation in this normalized frame. At the coarser level (level 1), 
we evaluate a uniform $20^3=8000$ grid over the $x$, $y$, and $z$ range $[-0.5, 0.5]$ (in normalized coordinates). For finer levels (levels 2-4), we perform coarse-to-fine refinement around the best unnormalized translation using $5$ samples per dimension, 
with level-dependent spacing $\tau_\ell \in \{0.05, 0.01, 0.005\}\,\mathrm{m}$. 
The most computationally intensive step occurs at the coarsest stage, 
where global $\SE(3)$ jointly samples $1024$ rotation hypotheses and $8000$ 
translation hypotheses, resulting in approximately $8 \times 10^6$ candidates. 
%, implemented via strided sampling followed by local densification. 
We use the same hierarchical search schedule for all the experiments.

\subsection{Non-Penetration Score (NPS)}
\label{app:nps}
As mentioned in Section \ref{sec:method}, three types of penetration are considered: inter-object penetration, object-environment penetration (e.g., table/ground), object-free-space penetration. For each object $i\in[N]$, let $\mathcal{J}(i)=\{0\}\cup\big([N]\setminus\{i\}\big)\cup\{\mathrm{free}\}$ denote the set of all surrounding entities for object $i$, where $0$ represents the environment and $\mathrm{free}$ represents free space. Let $p(i,j)$ return the deepest penetration between object $i \in [N]$ and $j\in \mathcal{J}(i)$:
\begin{equation}
\label{eq:penetration_depth}
p(i,j) \;=\;
\begin{cases}
\operatorname{ReLU}\!\Big(
-\min\Big(
\min\limits_{\bm{b}\in\mathcal{B}_j}\Phi_i\!\big(\bm{T}_i\!\cdot\bm{T}_j^{-1}\!\cdot\bm{b}\big),\; \\
\qquad \min\limits_{\bm{b}\in\mathcal{B}_i}\Phi_j\!\big(\bm{T}_j\!\cdot\bm{T}_i^{-1}\!\cdot\bm{b}\big)
\Big)
\Big), &\!\!\!\!\!\!\!\! j\in [N],\\[6pt]
\operatorname{ReLU}\!\Big(
-\min\limits_{\bm{b}\in\mathcal{B}_i}\Phi_{\mathrm{env}}\!\big(\bm{T}_i^{-1}\!\cdot\bm{b}\big)
\Big), &\!\!\!\!\!\!\!\! j=0,\\[6pt]
\operatorname{ReLU}\!\Big(
-\min\limits_{\bm{b}\in\mathcal{B}_i}\Phi_{\mathrm{free}}\!\big(\bm{T}_i^{-1}\!\cdot\bm{b}\big)
\Big), &\!\!\!\!\!\!\!\! j=\mathrm{free}.
\end{cases}
\end{equation}
Following \cite{zhang22icra-non}, the Non-Penetration Score (NPS) for object $i$ is
\begin{equation}
S^{NPS-ND}_{i}
:= \frac{1}{|\mathcal{J}(i)|}\sum_{j\in \mathcal{J}(i)} p(i, j),
\end{equation}
which evaluates the mean penetration across all objects in the scene.
This definition works well as a score for object $i$ if the shape and pose of each object $j$ is accurately estimated. 
% For example, object-free-space penetration is reliable if depth sensing is accurate. 
However, a single neighboring object with inaccurate shape or pose can significantly inflate the penetration metric for an accurately estimated object $i$.
% However, an inaccurate shape and pose estimation of a neighboring object can inflate the penetration metric of an accurate shape and pose estimation of object $i$. 
To mitigate this issue, we propose an adaptive alternative that excludes neighbors with large shape and pose errors:

\begin{equation}
S^{NPS}_{i}
:= \frac{1}{|\mathcal{J}(i)|}\sum_{j\in \mathcal{J}(i)} \mathbf{\mathcal{I}}(j) p(i, j),
\end{equation}
where $\mathbf{\mathcal{I}}(j)$ is an indicator function of whether object $j$ is included in the evaluation. For the experiment, we use ADD-S between as the indicator function 
because it encodes both shape and pose error of an object \cite{Xiang17rss-posecnn}.
We apply a threshold of 0.05 meters across all experiments. Additionally, due to noise in the depth maps which may affect free space estimation, we cap the maximum allowed penetration depth at 0.01 meters throughout all experiments.

\subsection{Additional Details for Picasso Dataset}
\label{app:dataset}
\begin{figure}[t!]
\centering
  \includegraphics[width=\columnwidth]{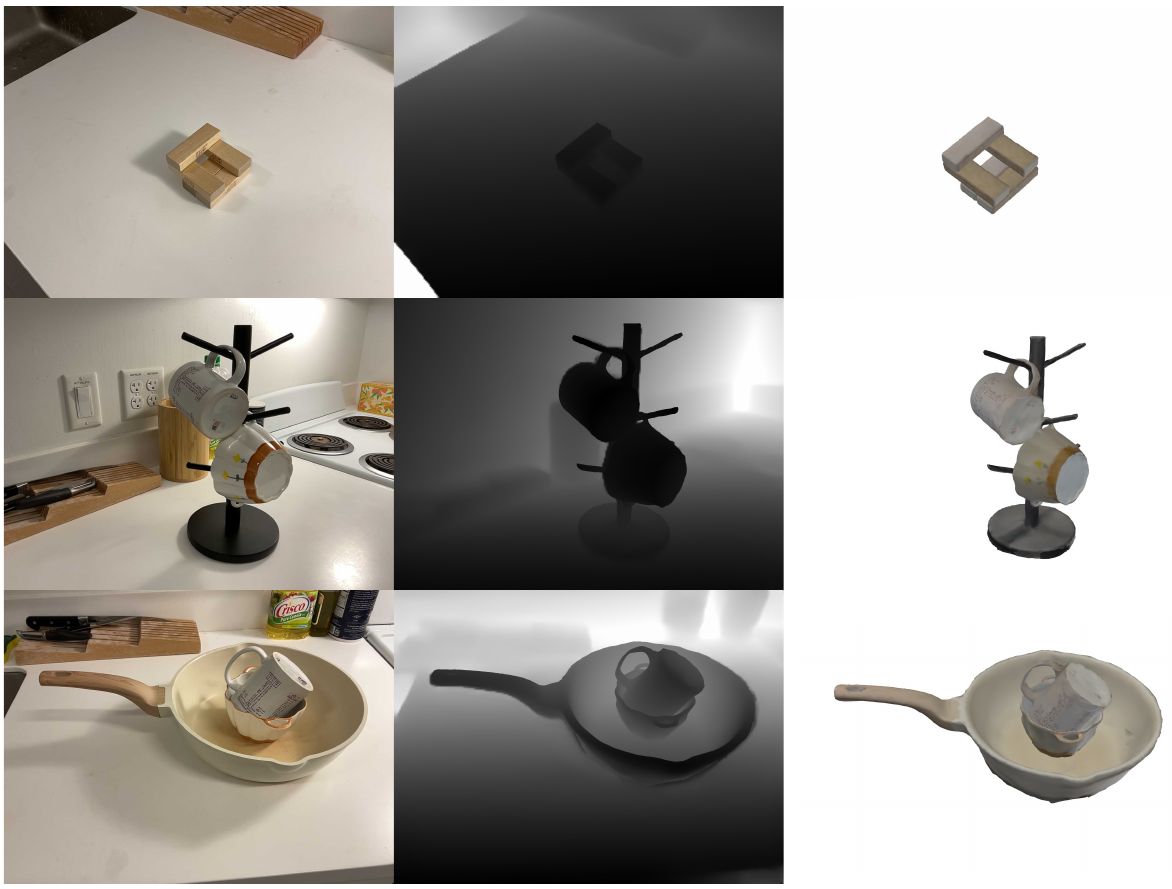}
\caption{Examples of Picasso Dataset. \textbf{Left: }RGB images. \textbf{Middle: }Depth maps. \textbf{Right: }Reconstructed 3D models.}
\label{fig:picasso_details}
\end{figure}

Fig.~\ref{fig:picasso_details} shows additional examples from the Picasso dataset. 
Picasso contains 10 contact-rich static scenes with 10 commonplace objects.

\subsection{Ablation Study}
\label{app:ablation}

Tables \ref{tab:ablation_study_1} and \ref{tab:ablation_study_2} show the ablation 
with physics constraints turned off on the YCB-V dataset in the CRISP experiments. 
The constraint-free version of Picasso achieves faster performance while full Picasso 
performs better both geometrically and in terms of physical plausibility.
We also evaluate constraint satisfaction rates.
Picasso with the physics corrector achieves satisfaction rates of
$72.34\%$, whereas the satisfaction rate drops to $52.25\%$ without
the physics correction.

% ================= Table 2 =================
\begin{table}[t]
\centering
\setlength{\tabcolsep}{6pt}
\caption{Evaluation results on the ADD-S ($\times 10^{-3}$ m) and ADD-S (AUC \%) metrics for the YCB-V dataset. CRISP-Syn+Picasso w/o phy: CRISP-Syn+Picasso with physics constraints turned off. \colorbox{green!60}{Best}.}
\label{tab:ablation_study_1}
\begin{tabular}{lccccc}
\hline
 & \multicolumn{2}{c}{ADD-S $\downarrow$} & \multicolumn{3}{c}{ADD-S (AUC \%) $\uparrow$} \\
\hline
Method & Mean & Median & 1 cm & 2 cm & 3 cm \\
\hline
\makecell[l]{CRISP-Syn\\+Picasso w/o phy}                  & {8.35}   & {3.04}   & {51.8} & {68.9} & {77.1} \\
\makecell[l]{CRISP-Syn\\+Picasso}            & \best{8.28}   & \best{2.98}   & \best{51.9}   & \best{69.1} & \best{77.2} \\
\hline
\end{tabular}
\vspace{-4mm}
\end{table}

% ================= Table 3 =================
\begin{table}[t]
\centering
\setlength{\tabcolsep}{6pt}
\caption{Evaluation results on the SPS, NPS ($\times 10^{-3}$ m), observable correctness (OC), and run time for the YCB-V dataset. CRISP-Syn+Picasso w/o phy: CRISP-Syn+Picasso with physics constraints turned off. \colorbox{green!60}{Best}.}
\label{tab:ablation_study_2}
\begin{tabular}{lcccc}
\hline
Method  & {SPS $\downarrow$} & {NPS $\downarrow$} & {OC $\uparrow$} & {run time (s) $\downarrow$} \\
\hline
\makecell[l]{CRISP-Syn\\+Picasso w/o phy}                   & {8.76}            & 4.06      & $54\%$  & \best{0.29}\\
\makecell[l]{CRISP-Syn\\+Picasso}               & \best{8.71} & \best{3.99} & \best{$55\%$} & 0.35      \\
\hline
\end{tabular}
\vspace{-2mm}
\end{table}

\begin{figure}[t]
\centering
  \includegraphics[width=\columnwidth]{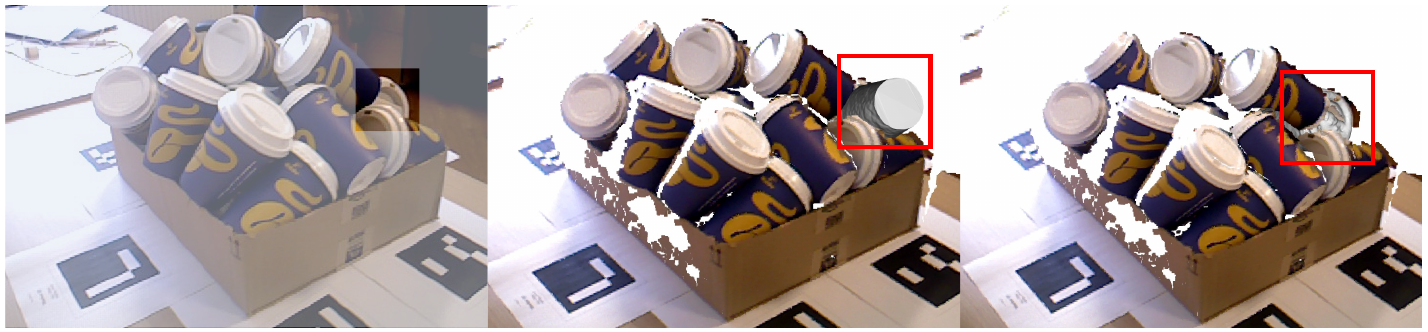}
\caption{
IC-BIN dataset experiment. \textbf{Left}: RGB image. \textbf{Middle}: w/o object-free-space constraints. \textbf{Right}: w/ object-free-space constraints.
}
\label{fig:icbin}
\end{figure}

Table~\ref{tab:ablation_each_term} shows the effect of each constraint term on Picasso. Inter-object and contact constraints
play a bigger role than rendering constraints like object-free-space. The effect of individual constraint terms appears to be
dataset-dependent, likely because their contributions depend
on visibility and ambiguity in the observed point clouds. On
YCB-V, the object-environment constraint produces the largest
performance gain, while the object-free-space constraint has
little measurable effect. In contrast, on IC-BIN, object-free-space constraints are more influential: when the observed point
cloud admits two plausible fits, the constraint helps reject the
solution that would violate free space as shown in Figure~\ref{fig:icbin}.

\begin{table}[t]
\centering
\caption{Ablation studies on YCB-V and IC-BIN datasets. \colorbox{green!60}{Best}.}
\label{tab:ablation_each_term}
\small
\setlength{\tabcolsep}{4pt}
\renewcommand{\arraystretch}{1.05}
\begin{tabular}{lc}
\hline
Method & OC $\uparrow$ \\
\hline
w/o object-environment & 54.42\% \\
w/o inter-object       & 55.76\% \\
w/o object-free-space  & 56.17\% \\
w/o contact            & 54.88\% \\
\hline
Picasso                & \best{56.17\%} \\
\hline
\multicolumn{2}{c}{\vspace{1pt}\normalsize YCB-V\vspace{1pt}} \\
\hline
Method & ADD-S (mm) $\downarrow$ \\
\hline
w/o object-environment & 8.58 \\
w/o inter-object       & 8.65 \\
w/o object-free-space  & 8.69 \\
w/o contact            & 8.59 \\
\hline
Picasso               & \best{8.58} \\
\hline
\multicolumn{2}{c}{\vspace{1pt}\normalsize IC-BIN\vspace{1pt}} \\
\end{tabular}
\vspace{-7mm}
\end{table}

\subsection{Hardware setup}
All experiments were performed on a machine with an NVIDIA RTX 4090 GPU.

\subsection{Boosting CRISP's Shape Prediction}

Tables \ref{tab:shape_correction_1} and \ref{tab:shape_correction_2} show the shape correction capability of Picasso with CRISP. CRISP uses latent shape code for implicit SDF field reconstruction. We select the top-k latent library codes that are have largest dot product with the predicted one and use Picasso to predict the object pose for each of them. Among these options we select the shape with lowest Chamfer loss. 
% In this way, we replace the latent shape code with the best performed shape code. 
We show in the experiments that this improves performance compared to CRISP alone in all metrics. Note that we give the mean value of Chamfer losses across all objects for shape error following \cite{Shi25cvpr-CRISP}.
This motivates future work on shape correction. For example, a generative model could produce multiple plausible shape candidates and we can extend sampling to the shape hypotheses.
\begin{table}
\centering
\setlength{\tabcolsep}{6pt}
\caption{Evaluation results on the $e_{\text{shape}}$ ($\times 10^{-3}$ m) and $e_{\text{shape}}$ (AUC \%) metrics for the YCB-V dataset. \colorbox{green!60}{Best}.}
\label{tab:shape_correction_1}
\begin{tabular}{lcccc}
\hline
\multirow{2}{*}{Method} & \multirow{2}{*}{$e_{\text{shape}}\downarrow$} & \multicolumn{3}{c}{$e_{\text{shape}}$ (AUC \%) $\uparrow$} \\
\cline{3-5}
 &  & 1 cm & 2 cm & 3 cm \\
\hline
CRISP-Syn+Picasso               & 26.95 & 47.6 & 63.3 & 77.4 \\
CRISP-Syn+Picasso (Shape)       & \best{26.90} & \best{47.8} & \best{63.4} & \best{77.4} \\
\hline
\end{tabular}
\end{table}

% ================= Table 3 =================
\begin{table}[t]
\centering
\setlength{\tabcolsep}{6pt}
\caption{Evaluation results on the physics metrics, penetration metrics, observable correctness, and runtime on the YCB-V dataset. \colorbox{green!60}{Best}.
%CRISP-Syn+P: CRISP-Syn with Picasso corrector.
}
\label{tab:shape_correction_2}
\begin{tabular}{lccc}
\hline
Method  & {SPS $\downarrow$} & {NPS $\downarrow$} & {OC $\uparrow$} \\
\hline
CRISP-Syn+Picasso               & {8.71} & {3.99} & {$55\%$}     \\
CRISP-Syn+Picasso (Shape)              & \best{8.64} & \best{3.74} & \best{$56\%$}  \\
\hline
\end{tabular}
\end{table}

\subsection{Physics-Guided Loss for Gradient Descent}
\label{app:physicsloss}

% Notation:
% - Objects j = 1..B, each has implicit SDF f_j(\cdot) in its own NOCS frame.
% - World point x \in R^3, homogeneous \bar{x} = [x;1] \in R^4.
% - Rigid transform T_{c\to j} maps world/cam coordinates into object-j NOCS coords.
% - SDF evaluated as s_j(x) := f_j( \pi( T_{c\to j}\,\bar{x} ) ), where \pi([p;w])=p \in R^3.
% - Sigmoid \sigma(z) = 1/(1+e^{-z}).

% -------------------------
% 1) Object–Object collision loss (loss_obj_obj)
% -------------------------

Physics-guided losses are widely used in soft-constrained and physics-informed machine learning \cite{malenicky25arxiv-physpose, yao25tg-cast}.
Focusing on multi-object non-penetration, we develop a suite of loss terms used by CRISP-Syn+GD 
baseline in Section \ref{sec:close_set}.
Recall that $\MT_i\in\SIM(3)$ maps points in the camera frame to the object frame.
Moreover, recall that $\Phi_i:\mathbb{R}^3\to\mathbb{R}$ denotes the signed distance field (SDF) of object $i$ in the local frame
and $\Phi_{\mathrm{free}}:\mathbb{R}^3\to\mathbb{R}$ denotes the free-space SDF.
We consider three types of penetration in this section:

\paragraph{Inter-object non-penetration}

\[
\mathcal{L}_{\text{inter-obj}}
=
\frac{1}{M^W}
\sum_{\vb \in \calB^W}
\operatorname{ReLU}\!\Big(
\sum_{i\in[N]}
\sigma\!\big(-\alpha\, \Phi_i(\MT_i \cdot \vb)\big)
-\tau
\Big),
\]
where $\sigma(z) = 1/(1+e^{-z})$ and $\calB^W=\{\vb_{j}\}_{j=1}^{M^W}\subset\mathbb{R}^3$ denote the camera frame object surface vertices 
in the overlapping axis-aligned bounding box regions across all objects. Empirically, we set \(\alpha=10000\), \(\tau=1.99\),
\(\lambda_{\text{io}}=10\). Intuitively, if an object penetrates with another $n$ objects, then $\sum_{i\in[N]}
\sigma\!\big(-\alpha\, \Phi_i(\MT_i \cdot \vb)\big)$ will be close to $n+1$.

\paragraph{Object-environment non-penetration}

\[
\mathcal{L}_{\text{env-obj}}
=
\frac{1}{M^E}
\sum_{\vb \in \calB^E}
\sum_{i\in[N]}
\operatorname{ReLU} \Big(-\Phi_i(\MT_i \cdot \vb)\Big),
\]
where $\calB^E=\{\vb_{j}\}_{j=1}^{M^E}\subset\mathbb{R}^3$ denote the camera frame sampled points below the plane.

% (Equivalent to your code: depth_beyond = clamp(-sdf - sdf_margin, 0),
%  then averaging over all collected penetrating points.)

% -------------------------
% 3) Free-space violation loss (loss_free_space)
% -------------------------
% For each frame k and object j:
% - Observed depth D_k(u) at pixel u=(p,q)
% - Rendered depth \hat{D}_{k,j}(u)
% Violation if \hat{D}_{k,j}(u) < D_k(u) - m, where m = free_space_depth_margin.
% Loss uses an exponential penalty on the positive depth difference.

\paragraph{Object-free-space non-penetration}

\[
\mathcal{L}_{\text{free}}
=
\frac{1}{M^F}
\sum_{b\in\calB^F}
\operatorname{ReLU} \Big(
e^{-\beta \Phi_{\mathrm{env}}(\vb)}
-1
\Big),
\]
where $\calB^F=\{\vb_{j}\}_{j=1}^{M^F}\subset\mathbb{R}^3$ denote the camera frame points lifted 
from pixel space and scaling factor $\beta=10$. Depth noise can cause false positives, so we use an exponential penalty that focuses on large violations. 
The total physics-guided loss is
\[
\mathcal{L}_{\text{phy}}
=
\lambda_{\text{inter-obj}} \mathcal{L}_{\text{inter-obj}}
+
\lambda_{\text{env-obj}} \mathcal{L}_{\text{env-obj}}
+
\lambda_{\text{free}} \mathcal{L}_{\text{free}},
\]
with $\lambda_{\text{inter-obj}}=10.0$, $\lambda_{\text{env-obj}}=1.0$ and $\lambda_{\text{free}}=10^{-10}$.

\subsection{Failure Cases}
\label{app:failure_cases}

\begin{figure}[t]
\centering
  \includegraphics[width=\columnwidth]{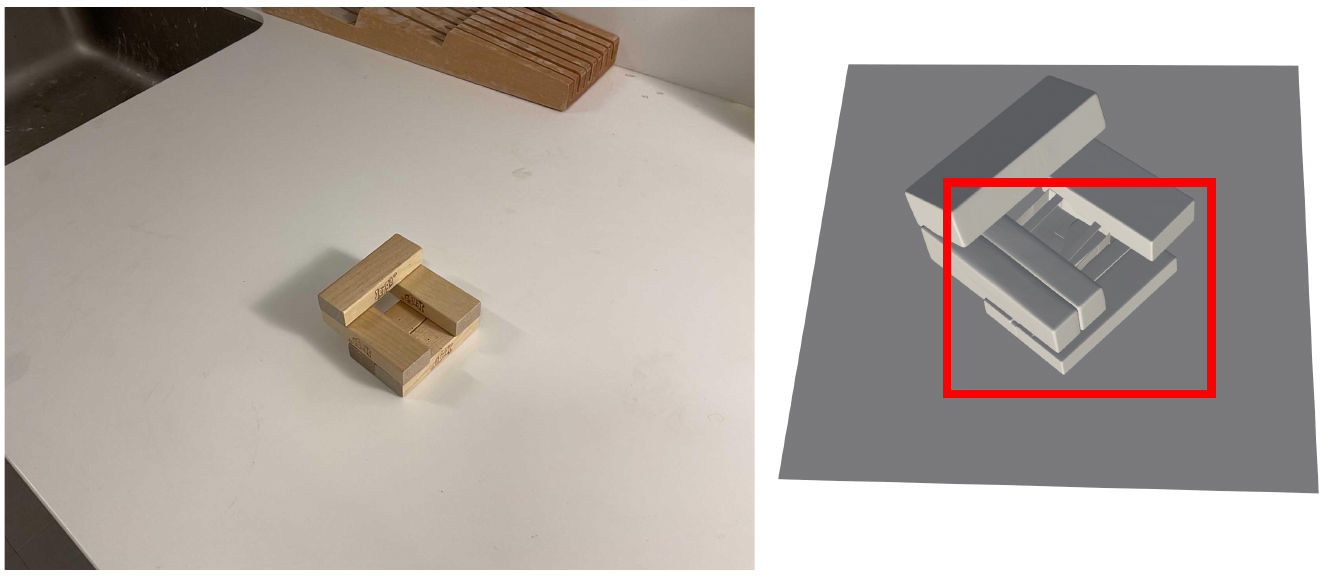}
\caption{A failure case due to inaccurate shape prediction.}
\label{fig:failure1}
\vspace{-2mm}
\end{figure}

\paragraph{Inaccurate Shape Prediction}
As shown in Fig.~\ref{fig:failure1}, inaccurate SAM3D shape predictions for the two bottom Jenga blocks lead to errors in Picasso's pose estimation.

\begin{figure}[t]
\centering
  \includegraphics[width=\columnwidth]{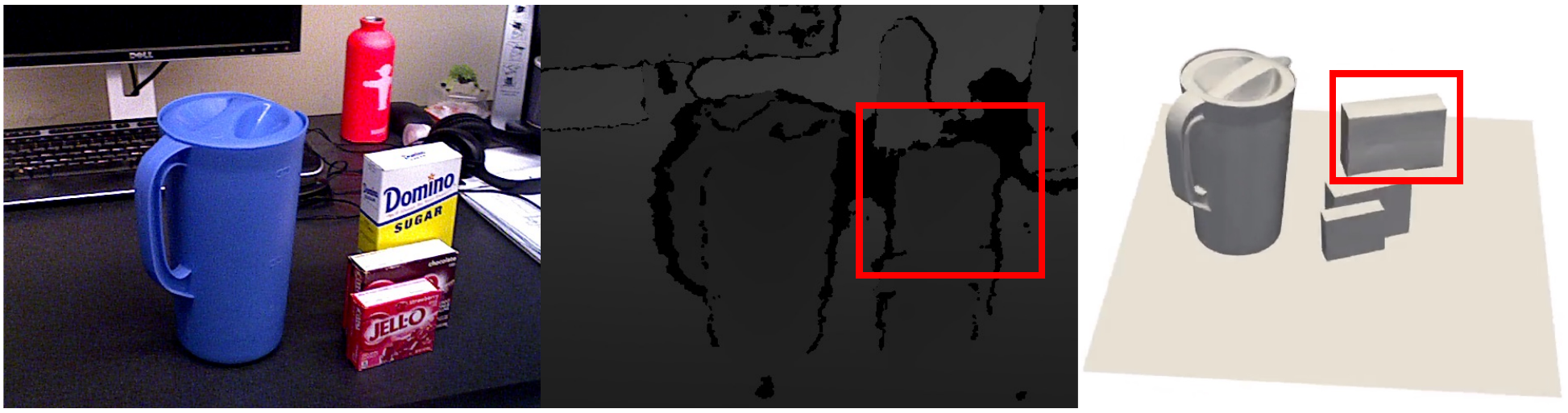}
\caption{A failure case due to noisy and partial depth point clouds.}
\label{fig:failure2}
\vspace{-2mm}
\end{figure}

\paragraph{Noisy or Partial Depth Point Clouds}
Noisy and partial depth point clouds also contribute to failures. In Fig.~\ref{fig:failure2}, 
the yellow box is only partially visible, yielding an incomplete depth point cloud. 
Picasso therefore selects an incorrect orientation that fits the observed points.
Violations of the physics (free-space) constraints in the final output occur 
because the correct configuration is not in the top-16 candidates considered for constraint satisfaction due to depth noise.

\paragraph{Contact Scene Graph Approximation}

\begin{figure}[t]
\centering
  \includegraphics[width=\columnwidth]{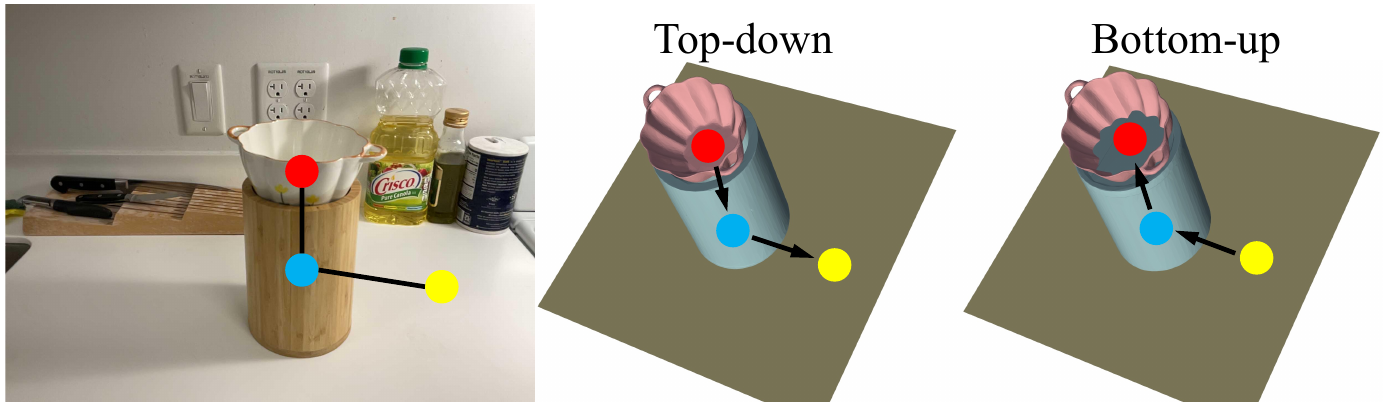}
\caption{Approximation of contact scene graph can lead to failure in downstream inference.}
\label{fig:failure3}
\vspace{-2mm}
\end{figure}

We perform a single sweep over a scene graph approximated as a DAG, with the ground node as the root. In some cases, this approximation can lead to incorrect predictions. As shown in Fig.~\ref{fig:failure3}, under a bottom-up DAG construction, the holder is inferred before the bowl, so non-penetration constraints between them are not enforced when predicting the holder. Due to occlusion, this results in the holder being estimated upside down. In contrast, with a top-down construction, the bowl is inferred first and the holder second, activating the non-penetration constraints and yielding the correct prediction.
This example motivates future work toward better inference schemes where upstream and downstream nodes in the contact scene graph mutually influence one another.

\subsection{Human Study: Alignment with Physics Metrics}
\label{sec:humansAppendix}

\textbf{Setup. } Our survey was performed between January 19th,  2026 and January 23rd, 2026. The survey involved two distinct groups: The first group was human participants from Amazon Mechanical Turk, which we refer to as the \emph{Public} group. We launched two batches of surveys (30 participants each), collecting 60 total responses. After removing duplicate submissions (as Mechanical Turk workers may complete the same survey multiple times), we obtained 59 valid responses. The second group consisted of researchers spanning robotics and computer vision, which we refer to as the \emph{Expert} group. To avoid bias toward any single research area, we included participants with expertise in 3D and low-level vision, control and motion planning, reinforcement learning, multi-modal learning, underwater robotics, SLAM, and rehabilitation robotics.

We selected three frames from each YCB-V sequence ---at the beginning, middle, and end--- yielding 36 images in total. For every image, we ran both SAM3D and SAM3D+Picasso to obtain the reconstructions. For each reconstruction, we created a video showing orbit views of the reconstruction.
Participants completed 36 trials. In each trial, they were shown two videos side by side: one generated by SAM3D and the other by SAM3D+Picasso. The video order was randomized per scene, and method names were anonymized to avoid bias. For both \emph{Expert} and \emph{Public} groups, participants received the following prompt: ``\emph{You will be given videos of a reconstructed scene. Imagine the scene is in simulation. Based on your intuition of the physical world ---we know that if two objects penetrate they will be pushed away in opposite direction, if an object is floating in the air, it will fall--- how would you evaluate the stability of the scene? Please evaluate from 1-7 where 1 means least stable and 7 means most stable.}''

\begin{figure}[t]
\centering
  \includegraphics[width=\columnwidth]{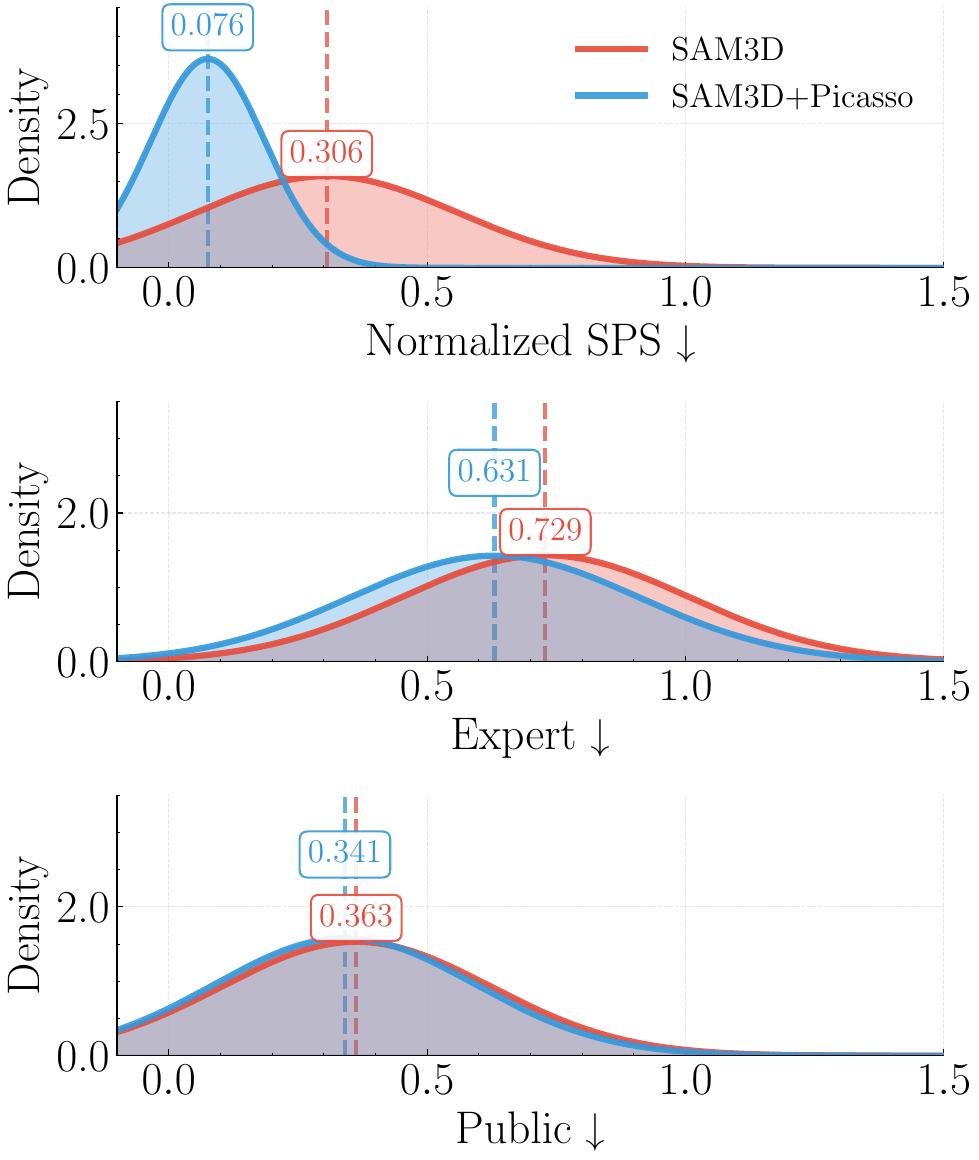}
\caption{Human evaluation and SPS comparison between SAM3D and SAM3D+Picasso. \textbf{Top: }SAM3D. \textbf{Bottom: }SAM3D+Picasso. For both \emph{Experts} and \emph{Public}, SAM3D+Picasso achieves higher physics plausibility.}
\label{fig:human_eval}
\vspace{-2mm}
\end{figure}

\myParagraph{Results} Fig.~\ref{fig:human_eval} compares human evaluation results between 
SAM3D and SAM3D+Picasso. We normalize both SPS and human evaluation scores, 
reversing the human evaluation metric so that lower values indicate better 
performance across all metrics. Both expert and public evaluators consistently rate 
SAM3D+Picasso as more physically plausible than SAM3D alone. 
While experts tend to assign lower scores overall 
(i.e., are more conservative in judging physical stability), 
the public is generally more optimistic. Moreover, 
the performance gap between SAM3D with and without the corrector is larger 
in the expert ratings, suggesting that experts are more sensitive to subtle 
physical inconsistencies. The performance gap between SAM3D with and without the corrector is 
even larger for SPS, indicating a better distinguishing capability of the metric.
A detailed per sequence results of SPS, expert and public data are shown in Table \ref{tab:human_eval}.
Overall, these results support that the corrector 
substantially improves the physical plausibility as for human intuition.

\begin{table}[t]
\centering
\caption{Human evaluation of physics plausibility and SPS across 12 YCB-Video trajectories. Human scores are on a 1-7 scale (higher is better) across 83 participants. SPS is averaged across 3 frames per trajectory. S: SAM3D. S+PC: SAM3D w/ physics-constrained corrector. \colorbox{green!60}{Best}.}
\label{tab:human_eval}
\resizebox{\columnwidth}{!}{%
\begin{tabular}{l|cc|cc|cc|cc}
\toprule
& \multicolumn{2}{c|}{\textbf{SPS} $\downarrow$} 
& \multicolumn{2}{c|}{\textbf{Expert} $\uparrow$} 
& \multicolumn{2}{c|}{\textbf{Public} $\uparrow$}
& \multicolumn{2}{c}{\textbf{Expert+Public} $\uparrow$} \\
\textbf{Trajectory} 
& {S\cite{chen25arxiv-sam}} & {S+PC} 
& {S\cite{chen25arxiv-sam}} & {S+PC} 
& {S\cite{chen25arxiv-sam}} & {S+PC} 
& {S\cite{chen25arxiv-sam}} & {S+PC} \\
\midrule
traj\_48 & 13.90 & \cellcolor{green!60}1.91 & 2.68 & \cellcolor{green!60}3.17 & 4.87 & \cellcolor{green!60}5.06 & 4.24 & \cellcolor{green!60}4.51 \\
traj\_49 & 3.05  & \cellcolor{green!60}0.54 & 2.32 & \cellcolor{green!60}2.88 & 4.76 & \cellcolor{green!60}4.88 & 4.04 & \cellcolor{green!60}4.29 \\
traj\_50 & 1.63  & \cellcolor{green!60}1.48 & 2.86 & \cellcolor{green!60}3.00 & 4.98 & \cellcolor{green!60}5.15 & 4.36 & \cellcolor{green!60}4.53 \\
traj\_51 & 7.75  & \cellcolor{green!60}0.65 & 2.78 & \cellcolor{green!60}3.03 & \cellcolor{green!60}5.07 & 4.94 & \cellcolor{green!60}4.41 & 4.38 \\
traj\_52 & 5.73  & \cellcolor{green!60}0.05 & 2.17 & \cellcolor{green!60}2.67 & 4.86 & \cellcolor{green!60}4.97 & 4.07 & \cellcolor{green!60}4.30 \\
traj\_53 & 7.66  & \cellcolor{green!60}0.00 & 2.39 & \cellcolor{green!60}3.74 & 4.69 & \cellcolor{green!60}4.99 & 4.01 & \cellcolor{green!60}4.62 \\
traj\_54 & 10.17 & \cellcolor{green!60}5.46 & 2.49 & \cellcolor{green!60}3.69 & 4.68 & \cellcolor{green!60}4.86 & 4.03 & \cellcolor{green!60}4.52 \\
traj\_55 & 6.20  & \cellcolor{green!60}2.49 & 2.21 & \cellcolor{green!60}3.72 & 4.61 & \cellcolor{green!60}5.07 & 3.90 & \cellcolor{green!60}4.68 \\
traj\_56 & 6.85  & \cellcolor{green!60}0.96 & 3.38 & \cellcolor{green!60}3.88 & 4.87 & \cellcolor{green!60}4.91 & 4.43 & \cellcolor{green!60}4.60 \\
traj\_57 & 2.08  & \cellcolor{green!60}0.06 & 2.44 & \cellcolor{green!60}3.53 & 4.78 & \cellcolor{green!60}4.97 & 4.10 & \cellcolor{green!60}4.54 \\
traj\_58 & 3.82  & \cellcolor{green!60}2.69 & \cellcolor{green!60}3.72 & 2.65 & \cellcolor{green!60}4.92 & 4.78 & \cellcolor{green!60}4.56 & 4.16 \\
traj\_59 & 4.58  & \cellcolor{green!60}1.84 & 2.08 & \cellcolor{green!60}2.65 & 4.80 & \cellcolor{green!60}4.87 & 4.01 & \cellcolor{green!60}4.22 \\
\midrule
\textbf{Overall} 
& 6.12 & \cellcolor{green!60}1.51
& 2.63 & \cellcolor{green!60}3.22
& 4.82 & \cellcolor{green!60}4.95
& 4.18 & \cellcolor{green!60}4.45 \\
\bottomrule
\end{tabular}%
}
\end{table}

% LS: each subfiguire should have titles: SAM3D, S + P.

%incorporate physics constraints directly into foundation-model training

\subsection{Hardware Demo}
\label{sec:hardware_exp}
A hardware demo is conducted to illustrate how Picasso can be used to build a digital twin for object-relation reasoning. 
We set up three Jenga blocks supporting a marker, and the task is to remove one block without disturbing the marker. 
We use the Waveshare RoArm-M2-S robotic arm and manually define end-effector waypoints based on object pose estimation for the manipulation policy. For qualitative results, please see the attached video.
% \begin{figure*}[t]
%   \centering
%   \includegraphics[width=\textwidth]{figures/hardware_demo.pdf}
%   \caption{\textbf{First Row: }SAM3D reconstruction is unstable. \textbf{Second through seventh rows: }Picasso's physics-based estimation yields a digital twin that is both geometrically and physically consistent, and that accurately predicts real-world behavior.}
%   \label{fig:hardware}
% \end{figure*}

\subsection{Runtime}
\label{sec:runtime}
We report the average runtime of key modules on the Picasso dataset. Picasso takes 4.15s per object with the physics corrector and 1.97s per object without it. A single SAM3D query takes 7.66s on average, while a single Gemini query for constructing the contact scene graph takes 11.71s.